\documentclass[preprint,12pt]{elsarticle}

\usepackage{amssymb}
\usepackage{amsmath}
\usepackage{pifont}
\newcommand{\cmark}{\ding{51}}
\newcommand{\xmark}{\ding{55}}
\usepackage{graphicx}
\usepackage{xspace}

\usepackage{multirow}
\usepackage{adjustbox}
\usepackage{makecell}
\usepackage{color}
\usepackage{soul}
\usepackage{bm}
\usepackage{enumitem}

\newcommand{\ctext}[3][RGB]{%
  \begingroup
  \definecolor{hlcolor}{#1}{#2}\sethlcolor{hlcolor}%
  \hl{#3}%
  \endgroup
}
\usepackage{tabularx}  
\usepackage{graphicx}
\usepackage{booktabs}
\usepackage[most]{tcolorbox}
\usepackage{mdframed} 
\usepackage{array} 
\usepackage{xcolor}  

\lstdefinestyle{datalogstyle}{
        basicstyle={\tt \scriptsize},  %
	xleftmargin={6pt},
        xrightmargin={6pt},
        columns=flexible,
        breakindent=0pt,
        breaklines=true, 
	frame=tb,
	stepnumber=1,
	firstnumber=1,
	numberfirstline=true,
	tabsize=2,
	extendedchars=true,
	breaklines=true,
	columns=fullflexible,
	keepspaces=true,
	escapeinside={@}{@},
	firstnumber=last,
	captionpos=b, 
	commentstyle=\color{black!65},
	numberstyle=\tiny\color{black!65},
	stringstyle=\color{codepurple},
	breakatwhitespace=false, 
	keepspaces=true,              
        mathescape=true, 
	numbersep=5pt,                  
	showspaces=false,                
	showstringspaces=false,
	showtabs=false,
	aboveskip={0.8\baselineskip},
	belowskip={0.2\baselineskip},
}
\definecolor{mygray}{RGB}{169,169,169}
\definecolor{myblue}{RGB}{0,102,204}
\definecolor{mygreen}{RGB}{0, 128, 0}
\definecolor{myred}{RGB}{255, 0, 0} 
\usepackage{colortbl}
\usepackage{url}
\usepackage{hyperref}
\usepackage{mathrsfs}

\usepackage{xcolor}
\usepackage{booktabs}
\usepackage{tabularx}
\usepackage{amssymb}
\usepackage{ragged2e}
\usepackage{geometry}
\usepackage{booktabs}

\usepackage[T1]{fontenc}
\usepackage{xspace}

\definecolor{darkgreen}{rgb}{0.0, 0.4, 0.13}

\newcommand\factscore{\textsc{PlainQAFact}\xspace}

\newcommand\benchmark{\textsc{PlainFact}\xspace}
\newcommand{\github}{\raisebox{-1.5pt}{\includegraphics[height=1.00em]{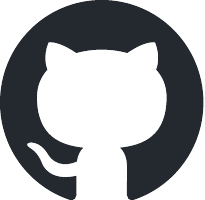}}\xspace}

\newif\ifrevision
\newif\ifrevisiontwo
\newif\ifrevisionthree

\revisionfalse    % Blue text will NOT be highlighted
\revisiontwofalse % Green text WILL be highlighted
\revisionthreefalse

\newcommand{\rev}[1]{%
  \ifrevision
    {\color{blue}#1}%
  \else
    #1%
  \fi
}

\newcommand{\orange}[1]{%
  \ifrevisiontwo
    {\color{orange}#1}%
  \else
    #1%
  \fi
}

\definecolor{myPurple}{HTML}{8A2BE2}

\newcommand{\purple}[1]{%
  \ifrevisionthree
    {\color{myPurple}#1}%
  \else
    #1%
  \fi
}

\begin{document}

\begin{frontmatter}



\title{\factscore: Retrieval-augmented Factual Consistency Evaluation Metric for Biomedical Plain Language Summarization}

\author[label1]{Zhiwen You} %% Author name
\author[label1]{Yue Guo}

\affiliation[label1]{organization={School of Information Sciences,\\ University of Illinois Urbana-Champaign},%Department and Organization
            city={\\Champaign},
            postcode={61820}, 
            state={IL},
            country={United States}}

%% Abstract
\begin{abstract}
Hallucinated outputs from large language models (LLMs) pose risks in the medical domain, especially for lay audiences making health-related decisions. Existing automatic factual consistency evaluation methods, such as entailment- and question-answering (QA) -based, struggle with plain language summarization (PLS) due to \textit{elaborative explanation} phenomenon, which introduces external content (e.g., definitions, background, examples) absent from the scientific abstract to enhance comprehension. To address this, we introduce \factscore, an automatic factual consistency evaluation metric trained on a fine-grained, human-annotated dataset \benchmark, for evaluating factual consistency of both source-simplified and elaborately explained sentences. \factscore first classifies sentence type, then applies a retrieval-augmented QA scoring method. Empirical results show that existing evaluation metrics fail to evaluate the factual consistency in PLS, especially for elaborative explanations, whereas \factscore consistently outperforms them across all evaluation settings. We further analyze \factscore's effectiveness across external knowledge sources, answer extraction strategies, answer overlap measures, and document granularity levels, refining its overall factual consistency assessment. 
Taken together, our work presents \rev{a sentence-aware, retrieval-augmented metric targeted at elaborative explanations in biomedical PLS tasks}, providing the community with both a \purple{new} benchmark and a practical \purple{evaluation} tool to advance reliable and safe plain language communication in the medical domain.
\factscore and \benchmark are available at: \github\url{https://github.com/zhiwenyou103/PlainQAFact}.

\end{abstract}

% \begin{figure*}[!ht]
%   \centering
%   \includegraphics[width=1.0\textwidth]{system.jpg}
% \end{figure*}

% \begin{highlights}
% \end{highlights}

%% Keywords

\end{frontmatter}

\section{Introduction}
Communicating biomedical scientific knowledge in plain language is essential for improving health information accessibility and health literacy~\cite{stoll2022plain, kuehne2015lay}. Recent advances in large language models (LLMs) have made significant progress in plain language summarization (PLS) of biomedical texts~\cite{ondov2022survey,Guo2023PersonalizedJI,goldsack2023domain,you-etal-2024-uiuc}. However, ensuring the factual consistency of these summaries remains a major challenge. A key source of inconsistency stems from \textit{elaborative explanations}: content such as definitions, background information, and examples that enhance comprehension but are not explicitly present in the original scientific abstracts (i.e., source) \cite{Guo2020AutomatedLL, srikanth2020elaborative, joseph-etal-2024-factpico}. While such elaborations are critical for effective communication, they introduce external content that cannot be directly verified against the source, complicating automatic factual consistency evaluation.

Factual consistency in PLS is typically assessed through a combination of human evaluation and automated metrics  \citep{ondov2022survey, jain2021summarization}. While human evaluation is reliable \cite{hardy2019highres}, it is costly and difficult to scale, particularly in biomedical contexts where domain expertise is required. Commonly used \purple{factual consistency evaluation} metrics can effectively verify content supported by the source but fail to assess factual consistency of added information \cite{guo-etal-2024-appls}. However, these metrics depend heavily on high-quality reference summaries, which are often unavailable in plain language summaries. Recent prompt-based evaluation techniques show promise \cite{luo2023chatgpt, you-etal-2024-beyond}, but their sensitivity to factual perturbations in elaborative content remains limited \cite{guo-etal-2024-appls}.

The lack of suitable benchmark datasets further hinders \purple{the progress of factual consistency evaluation.} Many existing datasets are constructed from LLM-generated summaries or apply rule-based perturbations to simulate non-factual content. For example, FactPICO provides expert annotations for plain language summaries of randomized controlled trial abstracts, focusing on PICO elements and evidence inference \cite{joseph-etal-2024-factpico}. However, it includes factuality labels only for added content, leaving simplified sentences unannotated, which are generated by LLMs and potentially inaccurate. In contrast, APPLS perturbs human-written summaries using rule-based transformations \cite{guo-etal-2024-appls}, but cannot ensure that the resulting outputs remain coherent or factually plausible. These limitations underscore the need for high-quality, sentence-level annotations grounded in human-authored summaries.

\begin{figure*}[!ht]
  \centering
  \includegraphics[width=0.98\textwidth]{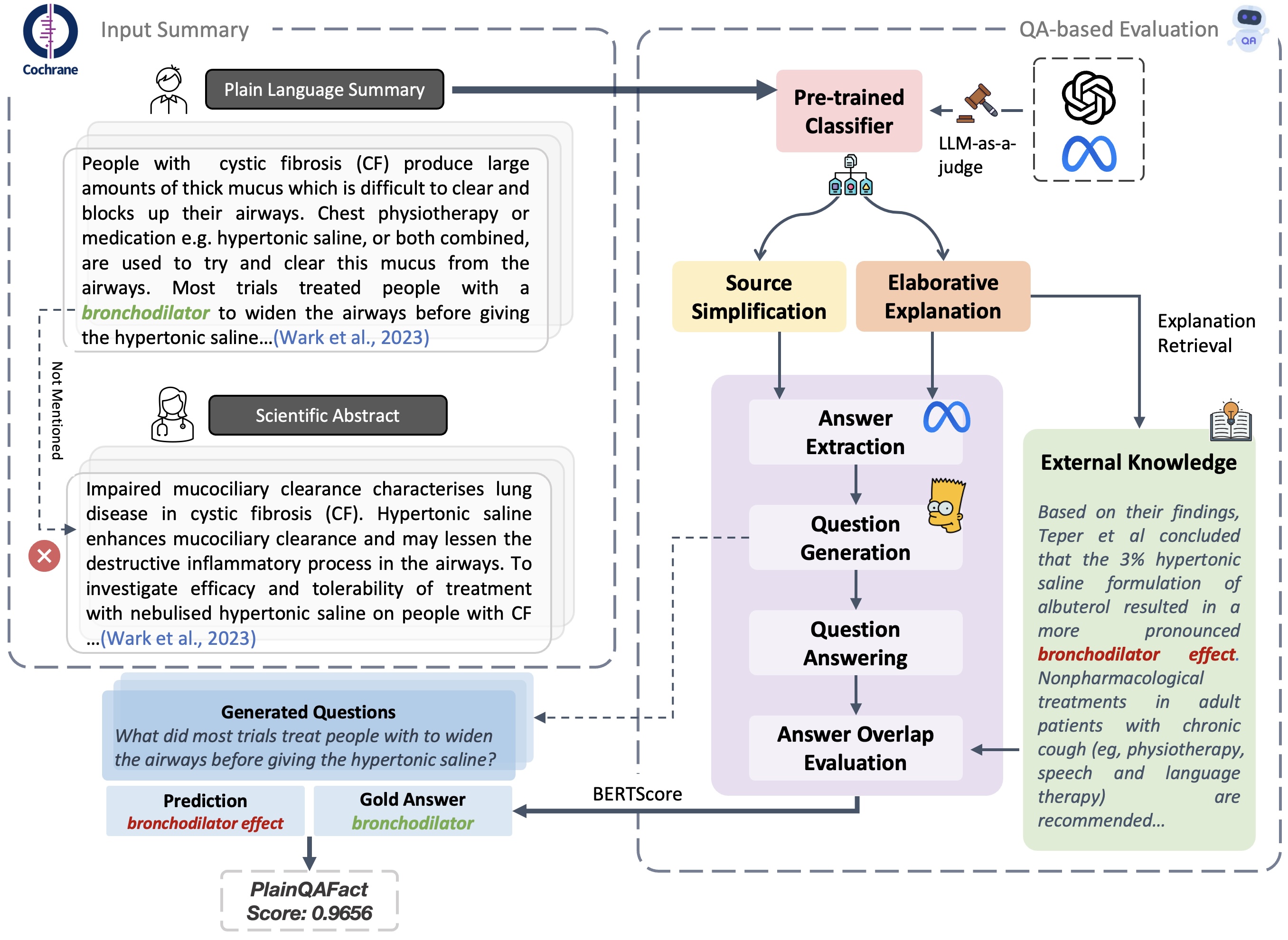}
  \caption{Overview of \factscore. A fine-tuned classifier first identifies the sentence type, involving either source simplification or elaborative explanation. Then, a QA-based evaluation pipeline performs answer extraction, question generation, question answering, and answer overlap evaluation. For elaborative content not present in the scientific abstract, \factscore retrieves external knowledge to verify factual consistency. The illustrated example shows an elaborative explanation involving a ``bronchodilator'' not mentioned in the source abstract but verifiable through external evidence. \factscore assigns a high score, reflecting strong alignment between the extracted and gold answers.}
  \label{fig:framework}
\end{figure*}

To address these challenges, we \purple{first} introduce a new expert-annotated dataset, \benchmark, consisting of human-written plain language summaries aligned with scientific abstracts. Each sentence is labeled with its type (elaborative explanation vs. source simplification), functional role, and alignment to the source (\S\ref{sec:benchmark}). Building on this \rev{and prior studies for the QA-based evaluation metrics in the general domain \cite{deutsch2021towards, fabbri2022qafacteval}}, \purple{we propose a targeted retrieval-based metric, \factscore, for factual consistency evaluation in PLS (Figure~\ref{fig:framework})}. It is a dual-stage QA-based evaluation metric that selectively applies retrieval only to elaborative explanations (\S\ref{methodology}). Experiment results on several PLS datasets demonstrate the effectiveness of our \purple{metric}, particularly in assessing complex, elaborative content (\S\ref{sec:experiments}, \S\ref{sec:results}). \purple{Taken together, our work introduces \purple{an automatic evaluation} metric specifically designed to assess factual consistency for PLS. In addition, we provide the community with a \purple{new} benchmark that supports nuanced evaluation of simplified and elaborative content. These contributions aim to promote more reliable and safe plain language communication in the medical domain.}

\section{Related Work}
\paragraph{Limitations of Existing Factuality Evaluation}
The primary approach for evaluating plain language generation combines automated metrics with human evaluation \citep{ondov2022survey, jain2021summarization}. While human assessment provides a thorough analysis \cite{hardy2019highres}, its high cost and time demands make it impractical for large-scale datasets. Evaluating factual consistency in biomedical PLS is particularly challenging, as it requires domain expertise. Entailment- \cite{lee2022factuality,laban2022summac}, similarity- \cite{wan2022evaluating,ye2024using}, model- \cite{zha2023alignscore}, and QA-based \cite{deutsch2021towards, fabbri2022qafacteval} metrics are commonly used for factual consistency assessment but rely heavily on high-quality reference summaries, which are often unavailable or difficult to obtain for PLS. Recent advancements in prompt-based evaluation show promise \citep{luo2023chatgpt}; however, their sensitivity to factuality perturbations in PLS remains limited \cite{guo-etal-2024-appls}. To address these limitations, we propose a reference-free solution for \purple{evaluating factual consistency in PLS, which effectively assesses factual consistency by leveraging external information retrieval to augment the reference summary.}

\paragraph{Retrieval-Augmented Generation}
Retrieval-augmented methods enhance text generation by extracting relevant information from external sources to supplement input queries \cite{lewis2020retrieval}. These methods have been shown to be effective in open-domain QA \cite{Ren2023InvestigatingTF, Mao2020GenerationAugmentedRF}, knowledge-based QA \cite{Kim2025MedBioLMOM}, and multi-step reasoning \cite{Gao2023RetrievalAugmentedGF, Tang2024MultiHopRAGBR}. In the context of PLS, retrieval from structured knowledge bases (KBs) has been shown to improve factual accuracy compared to language models alone \cite{guo2022cells}.
However, retrieval-augmented approaches have not been extensively explored for factual consistency evaluation in PLS, despite their potential for addressing elaborative explanations. In this work, we investigate retrieval-augmented QA \purple{scoring} to enhance factual consistency assessment in PLS while also examining its limitations.

% \section{\benchmark Benchmark} \label{sec:benchmark}

\section{Methods} 
We first introduce the protocol of \benchmark dataset curation (\S\ref{sec:benchmark}). Based on this dataset, we propose \factscore, a two-stage retrieval-augmented QA \purple{scoring} framework for factual consistency evaluation in PLS tasks (\S\ref{methodology}).

\subsection{\benchmark Benchmark} \label{sec:benchmark}

To develop a high-quality factual consistency evaluation benchmark in PLS tasks, we collect a subset from the largest human-authored CELLS \cite{guo2022cells} dataset (\S\ref{sec:benchmark_dataset}) and hire domain experts to provide fine-grained sentence-level annotations (\S\ref{sec:benchmark_annotation}).

\subsubsection{Human-Authored PLS Dataset} \label{sec:benchmark_dataset}
Rather than relying on LLM-generated plain language summaries, we construct our benchmark using human-authored summaries. CELLS \cite{guo2022cells}, the largest parallel corpus of scientific abstracts and their corresponding plain language summaries, is written by the original authors and sourced from 12 biomedical journals. We primarily select data from the Cochrane Database of Systematic Reviews (CDSR)\footnote{\url{http:///www.cochranelibrary.com}} within CELLS, as CDSR contains systematic reviews that support evidence-based medical decision-making across healthcare domains \cite{murad2016new}. Since systematic reviews represent the highest level of scientific evidence, this selection enhances the factual rigor of our dataset. To ensure readability, we filter the 200 most readable plain language summaries based on average scores from three standard readability metrics: Flesch-Kincaid Grade Level (FKGL) \cite{kincaid1975derivation}, Dale-Chall Readability Score (DCRS) \cite{dale1948formula}, and Coleman-Liau Index (CLI) \cite{coleman1975computer}. Additionally, given the collected plain language summaries from CELLS are all factual, we further conduct sentence-level perturbation to transform each plain language sentence into incorrect ones. Specifically, we use \texttt{GPT-4o}\footnote{We use \texttt{gpt-4o-2024-11-20} version for all experiments of \texttt{GPT-4o} in this paper.} to perturb each plain language sentence based on prompts introduced in \citet{guo-etal-2024-appls}. Therefore, for \benchmark, we have 400 summary-abstract pairs in total, where 200 factual pairs and 200 non-factual pairs.

\begin{table}[!ht]
\footnotesize
\begin{tabular*}{\textwidth}{@{\extracolsep\fill}lcc}
\toprule
                                 & Elaborative Explanation   & Source Simplification \\
\midrule
\# of Sentences                   &  1,213  &     1,527                     \\
Average Length (token)           &     29           &       28                  \\
Vocabulary Size                  &     4,230           &     4,046                    \\
Has Reference                &     417           &        1,527                 \\
\# of Background            &       533            &        329               \\
\# of Definition            &        82           &          44             \\
\# of Method/Result            &       512            &      1,107                 \\
\# of Example            &         10          &          3             \\
\# of Other            &          76         &          44             \\
\midrule
FKGL $\downarrow$  &                  12.5       &     12.4        \\
DCRS $\downarrow$&             11.3     & 11.6                \\
CLI $\downarrow$&             13.5     & 13.9                \\
\bottomrule
\end{tabular*}
\caption{Overview of the \benchmark benchmark. Medical experts annotated 200 pairs of plain language summaries and their corresponding scientific abstracts from the CELLS dataset \cite{guo2022cells}, categorizing each plain language sentence as either a source simplification or an elaborative explanation related to the abstract. Lower scores of FKGL, DCRS, and CLI indicate better readability.}
\label{tab:dataset_stats}
\end{table}

\subsubsection{Expert Annotation} \label{sec:benchmark_annotation}
Since each summary-abstract pair is authored by the same individual, we assume all information to be factual. The annotation aims to capture how plain language sentences relate to their scientific abstracts. 
Annotators analyze each plain language sentence across three dimensions: (1) \textbf{Factuality type}: identify whether a sentence is a source simplification (derived from the abstract) or an elaborative explanation (introducing new content); 
(2) \textbf{Functional role}: categorize the sentence as background, definition, example, method/result, or other; and (3) \textbf{Sentence alignment}: map each plain language sentence to its corresponding sentence(s) in the scientific abstract. Details of annotation guidelines are provided in~\ref{appx:annotation-protocol}. \orange{Note that ``functional role'' and ``sentence alignment'' are not components of developing \factscore. We use these two fields in \benchmark as metadata to conduct error analysis and document our annotation process in~\ref{appx:annotation-protocol} and ~\ref{appx:dataset-examples}.}

Annotations are conducted by four independent annotators, each with at least a bachelor's degree in biomedical sciences and prior fact-checking experience. Annotators are recruited via Upwork and compensated from \$15 to \$20 per hour. \rev{Each summary-abstract pair is annotated by two independent annotators, with disagreements resolved by a third. \orange{Inter-rater agreement (IRA), measured by Cohen’s Kappa for factuality type, functional role, and sentence alignment are 0.43, 0.60, and 0.55 respectively computed from 2,740 annotated plain language sentences, indicated fair agreement for all tasks \cite{artstein2008inter}. }}

Table~\ref{tab:dataset_stats} summarizes the dataset characteristics. Notably, 44\% of plain language sentences are elaborative explanations, highlighting their role in enhancing the readability of plain language summaries. 66\% of these cannot be directly verified against the source abstract, which underscores the need for factuality evaluation methods that account for such phenomena. Moreover, elaborative explanations include more background, definitions, and examples than source simplifications. For annotation examples, see~\ref{appx:dataset-examples}. Additionally, we provide qualitative analysis of \benchmark in ~\ref{appx:IAA-qualitative} given the fair IRA of factuality type.

\subsection{\factscore Framework} \label{methodology}
\factscore conducts fine-grained factual consistency evaluation for plain language summaries by first segmenting each summary into sentences, classifying as either simplification or explanation, and retrieving tailored external knowledge within a retrieval-augmented QA \purple{scoring} framework. Figure~\ref{fig:framework} provides an overview of its three key components: sentence-level classification (\S\ref{sec:classifier}), domain knowledge retrieval (\S\ref{sec:MedRAG}), and dual-stage QA-based factual consistency evaluation (\S\ref{sec:QAevaluation}). 

\subsubsection{Learned Factuality Type Classifier} \label{sec:classifier}
As elaborative explanations are prevalent in plain language generation and existing metrics struggle to capture added information \cite{guo-etal-2024-appls}, we first fine-tune a pre-trained language model to categorize factuality types of plain language sentences. Based on the factuality type annotations (source simplification vs. elaborative explanation) in \benchmark, we fine-tune the \texttt{PubMedBERT-base} model\footnote{\url{https://huggingface.co/microsoft/BiomedNLP-BiomedBERT-base-uncased-abstract-fulltext}} as a classifier. Our \benchmark is split 8:1:1 for training, validation, and testing. Additionally, we compare the pre-trained classifier with GPT-4o \cite{gpt4o} as a zero-shot classifier to explore extending our factual consistency evaluation metric to domains and tasks lacking human-annotated data.

\subsubsection{Domain Knowledge Retrieval} \label{sec:MedRAG}
Retrieval-augmented methods have proven effective for explanation generation \cite{guo2022cells, xu2024retrieval}, making them a natural fit for evaluating elaborative explanations. Since our dataset is in the medical domain, we employ MedCPT \cite{jin2023medcpt}, a retriever pre-trained on large-scale PubMed search logs to generate biomedical text embeddings. \rev{For external courpus, we incorporate StatPearls\footnote{\url{https://huggingface.co/datasets/MedRAG/statpearls}} \cite{xiong2024benchmarking} for clinical decision support and medical Textbooks\footnote{\url{https://huggingface.co/datasets/MedRAG/textbooks}} \cite{jin2021disease} for domain-specific medical knowledge (see details in ~\ref{appx:retrieval-statement}.}

\subsubsection{QA Evaluation Components} \label{sec:QAevaluation}
QA-based \purple{scoring} metrics have proved to be effective than other factual consistency evaluation metrics \cite{guo-etal-2024-appls} in general summarization tasks and align more closely with human annotations  \cite{deutsch2021towards, fabbri-etal-2022-qafacteval}. In this study, we adopt a QA-based \purple{scoring} approach as a backbone of our evaluation metric to verify factual consistency, incorporating answer extraction, question generation, question filtering, question answering, and answer overlap calculation modules.

\paragraph{Gold Answer Extraction (AE)} \label{sec:answer-extraction}
The first step in QA-based factual consistency evaluation is to extract answer entities (keyphrases) from plain language summaries as gold answer, then verify factuality by comparing them with answers generated by a QA model for the same questions. If the generated answer is correct or relevant, the summary is considered factual. The previous study used PromptRank \cite{kong2023promptrank}, a keyphrase generation method based on the \texttt{T5} model. To compare answer extraction strategies, we use PromptRank as the baseline, and evaluate LLM-based extractors, including an open-source \texttt{Llama 3.1 8B Instruct} model (\texttt{Llama 3.1}) \cite{llama3-2024} and close-source \texttt{GPT-4o} (\texttt{GPT-4o}) \cite{gpt4o}. Prompts for \texttt{Llama 3.1} and \texttt{GPT-4o} are \purple{presented} in~\ref{appx:llm-prompt}. 

\paragraph{Question Generation (QG)} \label{sec:question-generation}
Given an input plain language summary, the QG model generates questions based on the extracted answers (\S\ref{sec:answer-extraction}) and the input plain language summary. Following previous studies \cite{deutsch2021towards, fabbri2022qafacteval}\footnote{These studies use allennlp, an open-source NLP research library built on PyTorch, which is no longer actively maintained.}, 
we fine-tune \texttt{BART-large} model on standard QG datasets, including SQuAD \cite{rajpurkar-etal-2016-squad} and QA2D \cite{demszky2018transforming}, for use as the QG model in our evaluation metric. The QG model will generate multiple relevant questions based on the input, which will increase the probability of verifying the factual consistency by answering these questions in the following stages.

\paragraph{Question Filtering (QF)}
Questions generated by the QG model (\S\ref{sec:question-generation}) may not always be answerable. For example, the QG model generates \purple{``\textit{What can occur over several days?},''} which is vague and hard to predict the correct answer. To prevent these unanswerable questions from impacting the final evaluation performance, we follow QAFactEval \cite{fabbri-etal-2022-qafacteval} and remove the unanswerable questions using a pre-trained \texttt{Electra-large} model \cite{clark2020electra}. During QF, the filtering model receives only the plain language summary and its corresponding questions. In the subsequent QA stage (\S\ref{sec:question-answering}), answers are extracted from the source for answer overlap evaluation, while QF only determines whether the questions can be answered by the plain language summary.

\paragraph{Question Answering} \label{sec:question-answering}
The QA model extracts answers to answer the filtered questions from the source document. To prevent hallucinated output, we use an extractive QA model, a pre-trained \texttt{Electra-large}, which was the best performing QA model in QAFactEval \cite{fabbri-etal-2022-qafacteval}.

\paragraph{Answer Overlap Evaluation (AOE)}
We evaluate the alignment between the gold and generated answers from the QA model using the baseline Learned Evaluation metric for Reading Comprehension (LERC) score \cite{chen-etal-2020-mocha} and BERTScore \cite{zhang2019bertscore}. The final step in \factscore is weighting the factual consistency scores for source simplification and elaborative explanation sentences.  
Specifically, the simplification score $s$ is computed using only abstracts as the source in QA, while the explanation score $e$ is calculated by incorporating both abstracts and retrieved knowledge as source for QA-based evaluation. \rev{$n$ is the number of sentences in a plain language summary, after we split the summary using Natural Language Toolkit (NLTK)\footnote{\url{https://www.nltk.org/}} and and classify each sentence with our fine-tuned classifier.}
\begin{equation}\label{eq1}
\factscore = \frac{s_{\text{Avg.}} \cdot n_{s} + e_{\text{Avg.}} \cdot n_{e}}{n_{s} + n_{e}},
\end{equation}
where $n_{s}$ is the number of simplification sentences, and $n_{e}$ is the number labeled as explanation. $s_{\text{Avg.}}$ and $e_{\text{Avg.}}$ denote the average \factscore scores computed for instances classified as simplification and explanation sentences, respectively.

\section{Experiments} \label{sec:experiments}
We conduct main experiments on three publicly available datasets, including \benchmark, CELLS \cite{guo2022cells}, and FareBio \cite{fang2024understanding}. To verify the effectiveness of our proposed \factscore on PLS tasks, we compare with five widely used factual consistency evaluation metrics and two LLM-based evaluators. 

\subsection{Datasets}
As introduced in Sec~\ref{sec:benchmark}, we create \benchmark with fine-grained sentence-level annotations regarding the factuality types. Additionally, to test the generalizability of our pre-trained classifier, we also collect another subset from the test set of CELLS dataset \cite{guo2022cells}, gathering 200 summary-abstract pairs. Similar to \benchmark (\S\ref{sec:benchmark_dataset}), we also conduct sentence-level perturbation for each plain language summary in our collected CELLS dataset. Therefore, we have 200 factual and 200 non-factual summary-abstract pairs for both \benchmark and CELLS datasets. We also compare \factscore on the FareBio dataset, which contains 175 plain language summaries generated by seven different LLMs, together with the source articles \cite{fang2024understanding}. Specifically, FareBio includes 25 full scientific articles, each summarized by seven LLMs, resulting 175 plain language summaries. \rev{Because FareBio does not include non-factual simplification sentences, we treat all simplification sentences as ``factual.'' For each original summary, we form new plain language summaries by aggregating all sentences annotated as factual and all sentences annotated as non factual separately. This produces 33 non factual and 174 factual summaries. We also identify explanation sentences that receive the ``factual hallucination'' label (i.e., they are not verifiable from the source but are factually correct based on external knowledge) and build 53 factual explanation only summaries from them. We use the same set of 33 non factual summaries for both the main experiment (Table~\ref{tab:main-results}) and the explanation only evaluation (Figure~\ref{fig:farebio-external-only}).}

\subsection{Experiment Settings}
The classifier used in \factscore is fine-tuned on \benchmark with a batch size of 32 for 10 epochs. We apply early stopping during training based on validation loss. The temperature is set to 0 for \texttt{GPT-4o} and 0.01 for \texttt{Llama 3.1}. The \texttt{Llama 3.1} model used in AE also uses a temperature of 0.01. For the QG model, we fine-tune \texttt{BART-large} with a batch size of 16, a learning rate of 3e-5, and two training epochs. The maximum input length is set to 512 for all models in \factscore. More details of prompts and model hyper-parameters for \texttt{GPT-4o} and \texttt{Llama 3.1} are provided in~\ref{appx:experiment-settings}.

\subsection{Existing Factuality Metrics}
To the best of our knowledge, no prior work has efficiently evaluated factuality metrics for detecting elaborative explanations in PLS tasks. Moreover, most factuality metrics are designed for general-domain applications, largely due to limited quality and annotation of existing PLS datasets.  
To address this gap, we incorporate the following metrics in our experiment, informed by prior work \cite{joseph-etal-2024-factpico}:
(1) \textbf{Dependency-Arc Entailment (DAE)} \cite{goyal2021annotating} is an entailment-based method that evaluates summary factuality by breaking it into smaller entailment tasks at the arc level. The model independently determines whether each arc's relationship is supported by the input. (2) \textbf{AlignScore} \cite{zha2023alignscore} is a model-based factuality metric using \texttt{RoBERTa} \cite{liu2019roberta}. It extracts claims from the summary and calculates the alignment scores with all the context chunks from the input document. The final score is an average of all highest alignment probabilities. (3) \textbf{SummaC} \cite{laban2022summac} is an NLI-based inconsistency detection method designed for summarization tasks. It segments documents into sentences and aggregates scores between sentence pairs using a trained NLI model. (4) \textbf{QAFactEval} \cite{fabbri-etal-2022-qafacteval} is a QA-based factuality evaluation metric that assesses summary consistency by generating questions from the source document and comparing the model-generated answers with expected answers. (5) \textbf{QuestEval} \cite{scialom-etal-2021-questeval} is a reference‐less evaluation metric employs a \texttt{T5} \cite{2020t5} model to generate questions from the source document and verifies whether the summary can correctly answer them.

\section{Results and Analysis} \label{sec:results}
We first conduct a pilot study on the FactPICO dataset \cite{joseph-etal-2024-factpico} to investigate the performance of existing factual consistency evaluation metrics on PLS tasks. Our pilot study (\ref{app:pilot-study}) reveals two major limitations. First, many LLM-generated plain language summaries in FactPICO contain incomplete or incoherent sentences, highlighting the need for a benchmark based on high-quality, human-authored plain language summaries. Second, existing automatic factual consistency metrics are not sensitive to evaluate the factuality of ``added information,'' underscoring the need for a more domain-related factual consistency evaluation metric, especially for those summaries with additional explanations that are not mentioned by the original source texts. To address these challenges, we introduce a new expert-annotated dataset, \benchmark, and propose \factscore, a QA-based, retrieval-augmented metric for evaluating factual consistency in plain language summaries.

In this section, we first benchmark \factscore against five widely used automatic factual consistency evaluation metrics on three PLS datasets (\S\ref{sec:main_result}). We then evaluate its performance in an explanation-only setting using \benchmark (\S\ref{sec:explaination_result}), followed by an ablation study to assess the contribution of each component (\S\ref{sec:ablation}). Next, we conduct an error analysis to examine failure cases and highlight directions for further improvement (\S\ref{sec:error_analysis}).

\begin{table}[!ht]
\footnotesize
% \color{blue}
\begin{tabular*}{\textwidth}{@{\extracolsep\fill}llccc}
\toprule
\textbf{Datasets} & \textbf{Metrics} & \textbf{Kendall's~$\tau$} & \textbf{Pearson} & \textbf{AUC (99\% CI)} \\
\midrule
% \multicolumn{4}{l}{\textbf{CELLS}} \\
\multirow{8}{*}{CELLS} &    Llama 3.1      & 57.5 & 42.1 & 83.4* [78.4, 88.0] \\
    &    GPT-4o     & 70.0 & 75.4 & 99.3 [98.6, 99.8] \\
    \cmidrule(lr){2-5}
    & QAFactEval & 61.6 & 70.6 & 93.5 [90.2, 96.3] \\
    &    QuestEval  & 23.2 & 28.3 & 66.4* [59.1, 73.3] \\
    &    SummaC     & 24.8 & 29.8 & 67.6* [60.3, 74.3] \\
    &   AlignScore & 56.0 & 66.2 & 89.6 [85.3, 93.3]\\
    &    DAE        & \cellcolor[HTML]{F4CCCC}14.4 & \cellcolor[HTML]{F4CCCC}14.4 & \cellcolor[HTML]{F4CCCC}55.0* [50.7, 59.6]\\
    &    \factscore & \cellcolor[HTML]{D4EAD1}62.1 & \cellcolor[HTML]{D4EAD1}75.2 & \cellcolor[HTML]{D4EAD1}93.8 [90.3, 96.7] \\
\midrule
% \multicolumn{4}{l}{\textbf{FareBio}} \\
\multirow{8}{*}{FareBio} &    Llama 3.1      & 40.5 & 38.9 & 83.0 [69.0, 94.0] \\
    &    GPT-4o     & 27.3 & 35.7 & 75.7 [59.7, 89.8] \\
    \cmidrule(lr){2-5}
    &  QAFactEval & 32.8 & 43.9 & 81.6 [69.2, 91.7]\\
    &    QuestEval  & \cellcolor[HTML]{D4EAD1}37.8 & 45.2 & \cellcolor[HTML]{D4EAD1}86.4 [78.0, 93.3]\\
    &    SummaC     & 33.8 & 32.1 & 82.5 [69.7, 92.9]\\
    &    AlignScore & 32.2 & \cellcolor[HTML]{D4EAD1}47.2 & 81.0 [65.9, 93.0]\\
    &    DAE        & \cellcolor[HTML]{F4CCCC}8.8 & \cellcolor[HTML]{F4CCCC}8.8 & \cellcolor[HTML]{F4CCCC}53.4 [47.6, 57.6]\\
    
    &    \factscore & 16.8 & 35.6 & 66.1 [48.8, 82.1]\\
\midrule
% \multicolumn{4}{l}{\textbf{}} \\
\multirow{8}{*}{\benchmark} &    Llama 3.1     & 60.7 & 52.6 & 85.3* [80.6, 89.7]\\
    &    GPT-4o     & 69.8 & 80.3 & 99.2 [98.4, 99.8]\\
    \cmidrule(lr){2-5}
    & QAFactEval & \cellcolor[HTML]{D4EAD1}65.8  & 79.3  & \cellcolor[HTML]{D4EAD1}96.4 [94.0, 98.4] \\
    &    QuestEval  & 28.4 & 34.7 & 70.1* [63.1, 76.5]\\
    &    SummaC     & 34.1 & 42.8 & 74.1* [67.6, 80.4]\\
    &   AlignScore & 60.2 & 72.6 & 92.5 [93.8, 98.4]\\
    &    DAE        & \cellcolor[HTML]{F4CCCC}5.3 & \cellcolor[HTML]{F4CCCC}5.3 & \cellcolor[HTML]{F4CCCC}51.7* [47.5, 55.9]\\
    
    &    \factscore & 65.7 & \cellcolor[HTML]{D4EAD1}81.3 & \cellcolor[HTML]{D4EAD1}96.4 [93.6, 98.5]\\
\bottomrule
\end{tabular*}
\caption{Evaluation results of automatic metrics on the CELLS \cite{guo2022cells}, \benchmark, and FareBio \cite{fang2024understanding} datasets. Results are evaluated using Kendall’s~$\tau$, Pearson correlation, and AUC-ROC (AUC), with eight metric scores compared against human-labeled factual consistency. The standard deviations (std.) over five runs are: 0.2 (\factscore), 0.5 (\texttt{Llama 3.1}), and 6.0 (\texttt{GPT-4o}) on \benchmark; 0.1 (\factscore), 0.2 (\texttt{Llama 3.1}), and 6.3 (\texttt{GPT-4o}) on CELLS; and 0.1 (\factscore), 2.9 (\texttt{Llama 3.1}), and 17.0 (\texttt{GPT-4o}) on FareBio. * indicates a statistically significant difference compared to \factscore ($p < 0.01$). \textbf{Note the plain language summaries in FareBio are generated by seven LLMs, while CELLS and \benchmark are written by the authors of original articles.}}
\label{tab:main-results}
\end{table}

\subsection{Main Results} \label{sec:main_result}

We report rank correlation (Kendall's~$\tau$), linear correlation (Pearson), and discrimination (AUC-ROC \cite{bradley1997use}) for three datasets (\benchmark, CELLS~\cite{guo2022cells}, and FareBio~\cite{fang2024understanding}) following previous studies \cite{laban2022summac, fabbri-etal-2022-qafacteval, liu2019roberta} in Table~\ref{tab:main-results}. Overall, \texttt{GPT-4o} performs the best on CELLS and \benchmark, but its performance drops on FareBio. DAE is consistently the weakest across all datasets and metrics, suggesting that a dependency-only metric is insufficient for capturing the factual consistency required in PLS tasks.

On CELLS, \rev{\factscore outperforms all other automatic metrics on all three criteria. Specifically, 
\factscore achieves the best Kendall's~$\tau$ ($62.1$) and Pearson ($75.2$) among all the automatic factual consistency evaluation metrics. As a comparative effect size, the pairwise difference in AUC-ROC between \factscore and QAFactEval is small ($\Delta\mathrm{AUC-ROC}=0.3$, where $\Delta$ is \factscore minus QAFactEval). After Holm–Bonferroni correction \cite{holm1979simple} on AUC-ROC, \factscore shows significantly higher AUC-ROC than \texttt{Llama 3.1}, QuestEval, SummaC, DAE, and PromptRank ($p < 0.01$), whereas the differences with QAFactEval, AlignScore, and \texttt{GPT-4o} are not statistically significant (adjusted $p \ge 0.91$). This shows that our approach, first detecting elaborative content and then verifying it with retrieval, performs better than most of the existing reference-based metrics, which ask all questions from the source only. Similarly, on \benchmark, \factscore achieves the highest linear correlation with human ratings among automatic metrics (Pearson= $81.3$, 99\% CI [76.3, 85.4]), surpassing \texttt{GPT-4o}. It ties AUC-ROC (96.4) score with QAFactEval and has a competitive Kendall's~$\tau$ (65.7). As a comparative effect size, the pairwise difference in AUC-ROC between \factscore and QAFactEval is negligible ($\Delta\mathrm{AUC-ROC}=-0.1$). These results on both datasets show that \factscore is reliable for evaluating the factual consistency of plain language summaries. See ~\ref{appx:stat-tests} for more details of statistical tests.}

On FareBio, no single method is best across all criteria. \rev{QuestEval achieves the best Kendall's~$\tau$ ($37.8$) and AUC-ROC ($86.4$), while AlignScore achieves the best Pearson correlation ($47.2$).} However, FareBio differs from the other two datasets in several important ways: \rev{(1) the original sentence labels in FareBio are incomplete. Any sentence that can be traced back to the source is labeled as factual, so the dataset does not include sentences labeled as non-factual simplifications. This design leads to unreliable final summary-level labels. For example, even if a summary is labeled as factual and all of its sentences are labeled as factual, some sentences may still contain factual errors; (2) the processed dataset is highly imbalanced (33 non-factual vs.\ 174 factual summaries); (3) FareBio reports a fair Cohen’s Kappa of $0.48$ \orange{computed from 34 plain language sentences} without adjudication, which indicates notable annotation quality issues in their curated dataset; and (4) all plain language summaries are generated by LLMs, and the plain language sentences are not guaranteed to be correct, since the annotation only checks whether the sentence is a source simplification without post-processed non-factual perturbation. These factors likely increase the difficulty of evaluation on FareBio and affect the performance of factual consistency evaluation metrics, including \factscore and \texttt{GPT-4o}.} 

\rev{We also provide a detailed error analysis of FareBio that explains the main limitations of both \factscore and the dataset (\S\ref{error-analysis-farebio}). In addition, because the non-factual plain language summaries of \benchmark and CELLS are generated by \texttt{GPT-4o} (\S\ref{sec:benchmark_dataset}), they may contain detectable artifacts that are not representative of real PLS errors. To further test the robustness of our proposed metric, we conduct additional experiments using the heuristic entity swap approach introduced in \citet{guo-etal-2024-appls} to create non-factual plain language summaries. Specifically, following the same experiment setting as APPLS \cite{guo-etal-2024-appls}, we replace entities in each plain language sentence using the KBIN method \cite{wright2022generating}, which replaces specific entity spans (such as names of viruses or diseases) with related but different concepts derived from the Unified Medical Language System (UMLS)\footnote{\url{https://www.nlm.nih.gov/research/umls/}} while maximizing NLI contradiction and minimizing LM perplexity. This approach ensures that each new perturbed sentence is not only different from the original sentence, but also logically contradicts it while remaining grammatically fluent. Overall, on the CELLS dataset, for both the LLM-perturbed setting and the heuristic entity swap method, our metric achieves the best performance across all three evaluation criteria compared with other automatic metrics, showing the robustness of our proposed metric. Detailed experiment results are reported in ~\ref{appx:entity-swap}.}

In summary, \factscore is a more consistent and effective metric for factual consistency \purple{evaluation} compared to existing automatic \purple{metrics}. It clearly improves over prior QA-based metrics on CELLS and matches or exceeds them on \benchmark. Although \texttt{GPT-4o} performs better in most of the settings, our \factscore is built entirely on the open-source LLM (i.e., \texttt{Llama 3.1}), ensuring transparency, reproducibility, and accessibility. Its two-step evaluation process, detecting explanations and verifying them with retrieval, helps align metric scores more closely with human judgments in PLS evaluation tasks.

\begin{figure*}[!h]
  \centering
  \includegraphics[width=0.9\textwidth]{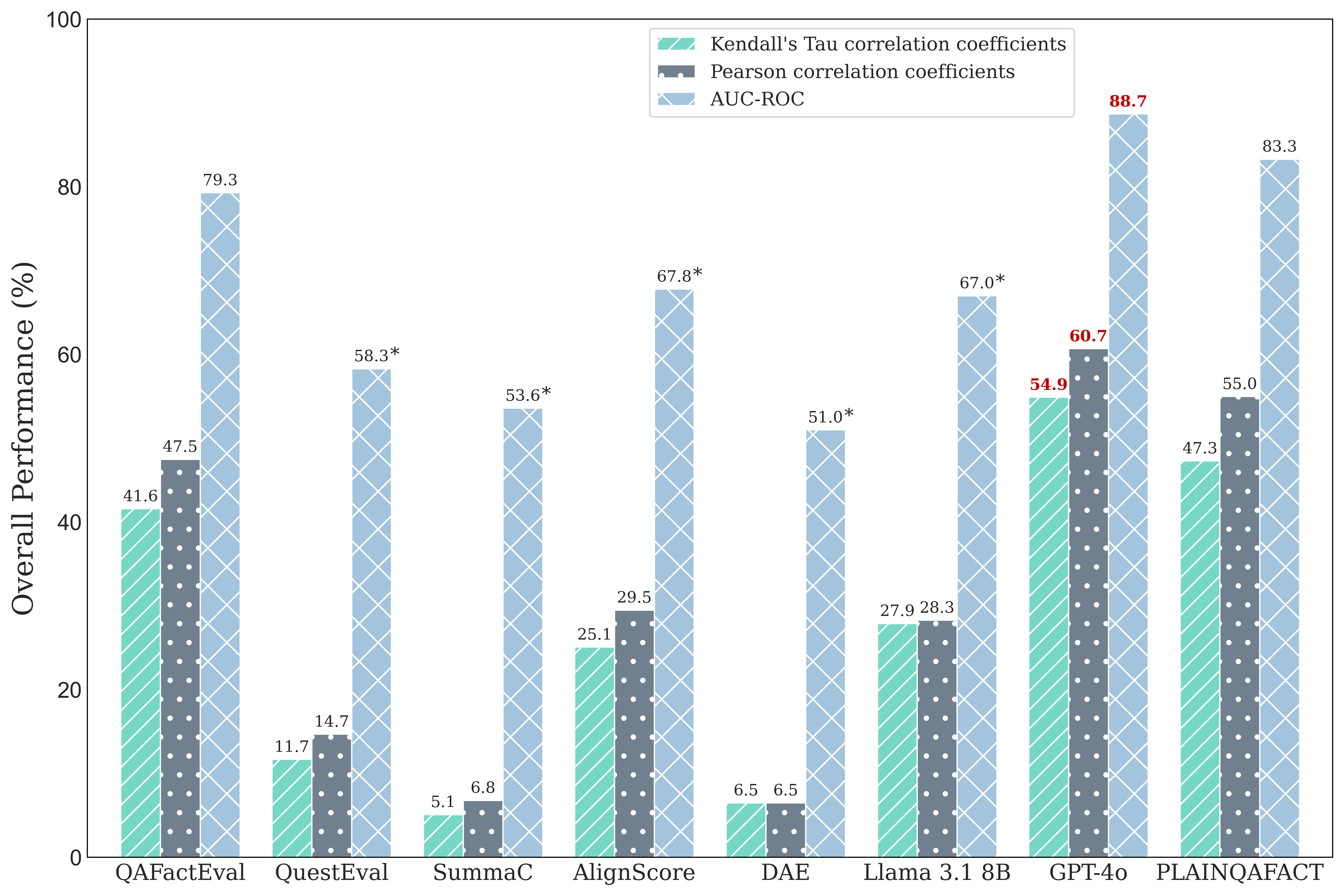}
  \caption{Overall performance on human-annotated elaborative explanation summaries from \benchmark (392 summaries). The std. of \factscore, \texttt{Llama 3.1}, and \texttt{GPT-4o} are 0.1, 1.0, and 7.7, respectively based on five runs for each metric. * indicates a statistically significant difference compared to \factscore ($p < 0.01$). \factscore significantly outperforms most of the automatic factual consistency evaluation metrics in AUC-ROC. Note that the CELLS dataset does not contain annotations for elaborative explanations. Results of explanation-only evaluation on FactPICO and FareBio are reported in~\ref{app:explanation-only}.}
  \label{fig:external-only}
\end{figure*}

\subsection{Explanation-Only Evaluation} \label{sec:explaination_result}

To test whether existing evaluation metrics are limited in assessing elaborative explanations in PLS, we evaluate \factscore on human-annotated explanation-only summaries from \benchmark. We select only the sentences labeled as ``elaborative explanation'' and group them into plain language summaries, resulting in 392 summary–abstract pairs for \benchmark. Note that CELLS does not provide sentence-level annotations for elaborative explanations, so we evaluate it only as a general PLS dataset (\S\ref{sec:main_result}).

As shown in Figure~\ref{fig:external-only}, \rev{\factscore achieves robust performance with an AUC-ROC of $83.3$ ($95$\% CI [$77.7$, $88.4$]) and a Pearson correlation of $55.1$, outperforming most automatic baseline metrics and approaching \texttt{GPT-4o} (AUC-ROC$=88.7$, $99$\% CI [$84.2$, $92.8$]). Comparison with open-source metrics, \factscore demonstrates a notable improvement over standard open-source, QA- and NLI-based metrics. It produces statistically significant improvements ($p<0.01$) over \texttt{Llama 3.1}, AlignScore, SummaC, QuestEval, and DAE across all correlation measures. For instance, \factscore improves upon the zero-shot \texttt{Llama 3.1} baseline by a margin of $\Delta$AUC-ROC$=16.3$, showing that instruction-tuned LLM judges alone are not enough for reliable evaluation of elaborative content. Comparing with the state-of-the-art QA-based metric QAFactEval, \factscore achieves a higher absolute AUC-ROC ($\Delta\mathrm{AUC-ROC}=4.0$) and Pearson correlation (Pearson$=55.1$). However, after applying the Holm-Bonferroni correction, this improvement is on the margin of statistical significance ($p=0.056$). These results indicate that \factscore captures the factual consistency of elaborative explanations more effectively than most of the prior automatic metrics. While \texttt{GPT-4o} achieves higher overall scores in all three criteria, it shows much higher variance (std.\ 7.7) compared to \factscore (std.\ 0.1), making it less stable for factual consistency evaluation.}

Compared to the results in Table~\ref{tab:main-results}, AlignScore is competitive on general PLS tasks but drops on evaluating explanation-only sentences. In contrast, \factscore remains consistently strong when we isolate explanation-only content. \factscore detects added elaborative explanations and checks them with domain knowledge retrieval, which helps when factual consistency depends on facts beyond simple restatement from the source. Since 44\% of sentences in our curated benchmark are labeled as elaborative \purple{explanations} by human annotators, we believe that \factscore is the more suitable and robust factual consistency evaluation metric in this scenario.

\begin{table}[!ht]
\footnotesize
% \color{orange}
\begin{tabular*}{\textwidth}{@{\extracolsep\fill}llcccc}
\toprule
\textbf{Component}                 & \textbf{Method Choice} & \textbf{$\tau$} & \textbf{Pearson} & \textbf{AUC (99\% CI)} & \textbf{std.} \\
\midrule
\textbf{\factscore} &                       & 65.7 & 81.3 & 96.4 [93.6, 98.5] & 0.2 \\
\midrule
\multirow{3}{*}{Classifier}            & \textbf{Fine-tuned classifier}    & -  & -   & -  & -   \\
                                   & GPT-4o        & 62.8 & 77.1   & 94.3* [90.8, 97.1] & 0.1   \\
                                   & No (retrieve for all) & 61.2  & 74.2  & 93.2* [89.6, 96.2] & 0.7  \\
\midrule
\multirow{3}{*}{AE}                  & \textbf{Llama 3.1} & -   & -   & -   & -  \\
                                   & GPT-4o               & 66.9 & 81.9   & 97.2 [95.2, 98.9] & 0.6   \\
                                   & PromptRank            & 66.3 &  80.2  &  96.0 [94.2, 98.4] &  -  \\
\midrule
\multirow{3}{*}{Retrieval Source}                   & \textbf{\textsc{Abs + TB + SP}} & -   & -   & -   & -   \\
& TB + SP & 63.4  & 76.8   & 94.8* [91.7, 97.3] & 0.1      \\
                                   & \textsc{Abs} (no retrieval)              & 62.8 & 77.1   & 94.3* [90.8, 97.1] & 0.1       \\
\midrule
\multirow{2}{*}{Answer Overlap}                     & \textbf{BERTScore}             & -    & -   & -   & -  \\
                                   & LERC                  & 64.5 & 75.9   & 95.6 [92.7, 97.9] & 0.1   \\
\midrule
\multirow{2}{*}{Granular Level}                      & \textbf{Sentence}             & -    & -   & -  & -   \\
                                   & Summary                  & 49.3  & 59.6   & 84.7* [79.6, 89.5] & 0.5   \\
\bottomrule
\end{tabular*}
\caption{Ablation study of \factscore on \benchmark, analyzing the impact of individual components in \factscore. $\tau$: Kendall's~$\tau$; AE: answer extraction; \textsc{Abs}: abstract; \textsc{TB}: Textbooks; \textsc{SP}: StatPearls; AUC: AUC-ROC. The first row of each component setting represents our best combination. * indicates a statistically significant difference compared to \factscore ($p < 0.01$). We run each setting for five times and report the standard deviation (std.). Note that only the settings using LLMs may produce fluctuated scores.}
\label{ablation_study}
\end{table}

% \begin{table}[!ht]
% \footnotesize
% \color{orange}
% \begin{tabular*}{\textwidth}{@{\extracolsep\fill}llcccc}
% \toprule
% \textbf{Component}                 & \textbf{Method Choice} & \textbf{$\tau$} & \textbf{Pearson} & \textbf{AUC (99\% CI)} & \textbf{std.} \\
% % \midrule
% % \textbf{\factscore} &                       & 65.7 & 81.3 & 96.4 [93.6, 98.5] & 0.2 \\
% \midrule
% \multirow{3}{*}{Classifier}            & \textbf{Fine-tuned classifier}    & 62.1  & 75.2   & 93.8 [90.3, 96.7]  & 0.1   \\
%                                    & GPT-4o        & 62.0 & 74.7   & 93.4 [90.1, 96.5] & 0.2  \\
%                                    & No (retrieve for all) & 60.6  & 73.4 & 92.8* [89.0, 95.9] & 0.5  \\
% \bottomrule
% \end{tabular*}
% \caption{Ablation study of \factscore on CELLS, analyzing the impact of individual components in \factscore.}
% \label{cells_ablation_study}
% \end{table}

\subsection{Ablation Study} \label{sec:ablation}
\factscore consists of several modules, including a fine-tuned classifier, QA modules, a retrieval function, and an answer overlap evaluation process. To understand the contribution of each component, we conduct an ablation study on \benchmark. Table~\ref{ablation_study} summarizes the ablation results, where each component of \factscore is individually modified to measure its impact on overall performance.

\noindent \textbf{Fine-tuned Classifier}
Table~\ref{ablation_study} shows that removing the classifier and retrieving for every sentence does not improve factual consistency assessment. \rev{\factscore statistically significantly outperforms the settings using \texttt{GPT-4o} as the classifier and the setting without a classifier \orange{($p < 0.01$)}}. Performance drops across all criteria (Kendall's~$\tau$ from 65.7 to 61.2; Pearson from 81.3 to 74.2; AUC-ROC from 96.4 to 93.2), and the std. increases to 0.7. This approach also requires notably more computation. Using \texttt{GPT-4o} as a simple classifier provides reasonable results but still underperforms compared to our fine-tuned classifier. These findings support the use of a lightweight fine-tuned classifier to trigger retrieval only when necessary.

\noindent \textbf{Answer Extraction}
\rev{For extracting gold answers from plain language summaries, the LLM based extractors perform better than PromptRank \cite{kong2023promptrank}. \texttt{GPT-4o} achieves the best overall scores (Kendall's~$\tau$=66.9, Pearson=81.9, AUC-ROC=97.2), with only small gains over our Llama-based method. In \benchmark, PromptRank obtains a slightly lower Pearson correlation than \factscore. However, in the ablation study on the CELLS dataset in Table~\ref{tab:answer-extractor-ablation}, \factscore significantly outperforms PromptRank in both perturbation settings (see details in ~\ref{appx:entity-swap}). Overall, we do not find a statistically significant difference between \factscore and the other answer extraction methods. \factscore is statistically tied with PromptRank (AUC-ROC=$96.0$, $p=0.34$) and with \texttt{GPT-4o} AE (AUC-ROC=$97.2$, $p=1.0$). Although \texttt{GPT-4o} achieves slightly higher performance overall, the marginal gain does not justify the increased cost, making \texttt{Llama 3.1} a more practical and cost-effective choice for large-scale factual consistency evaluation.}

\begin{table}[!ht]
\footnotesize
% \color{blue}
\begin{tabular*}{\textwidth}{@{\extracolsep\fill}llcccc}
\toprule
\textbf{Component}                 & \textbf{Method Choice} & \textbf{$\tau$} & \textbf{Pearson} & \textbf{AUC (99\% CI)} & \textbf{std.} \\
\midrule
Simplification         & \textsc{Abs}              &   63.5  & 79.0 & 94.8 [91.6, 97.4] &  0.2  \\
\midrule
\multirow{7}{*}{Explanation}     & Abs    &   32.8     &  38.2  & 72.7* [65.8, 79.2]  &  0.2  \\
                 & \textsc{TB}               &    31.2   & 37.2      &   74.6* [68.0, 80.9]    & 0.1   \\
                 & \textsc{SP}               &    33.3  &    37.8     &  72.5* [65.9, 79.1]    &  1.7  \\
                 & \textsc{TB + SP}          &    36.7     &     42.0     &      75.4* [68.8, 81.7]    &  0.1   \\
                 & \textsc{Abs + TB}           &    40.3   &     47.0    &   78.1* [71.8, 84.1]   & 0.2   \\
                 & \textsc{Abs + SP}           &    40.5  &    47.4   &   78.3* [71.7, 84.2]  &   0.2  \\
                 & \textsc{Abs + TB + SP}        &     \textbf{44.2}  & \textbf{51.3}  & \textbf{81.0} [74.9, 86.7]  &   0.1   \\
\midrule
\multirow{7}{*}{Full Dataset}              & \textsc{Abs}              & 62.8 & 77.1   & 94.3* [90.8, 97.1] & 0.1   \\
                 & \textsc{TB}                &    60.8   & 73.8  & 94.0* [90.7, 96.7]   &  0.1 \\
                 & \textsc{SP}                &    61.3   & 74.2   & 93.3* [89.6, 96.4]   &   0.1 \\
                 & \textsc{TB + SP}          &    63.4  &    76.8    &   94.8* [91.7, 97.3]    &  0.1  \\
                 & \textsc{Abs + TB}          &     64.3  & 79.7    &   95.4* [92.2, 97.9]    &   0.2 \\
                 & \textsc{Abs + SP}           &     65.0   &    80.3    &   95.9 [92.9, 98.2]   &   0.2\\
                 & \textsc{Abs + TB + SP}         &     \textbf{65.7} &   \textbf{81.3}    &   \textbf{96.4} [93.6, 98.5]    &   0.2 \\
\bottomrule
\end{tabular*}
\caption{Ablation study results on the \benchmark, evaluating how different retrieval sources affect various information types using \factscore. The fine-tuned classifier categorizes input sentences as \textbf{source simplification} or \textbf{elaborative explanation}. Overall, the numbers of factual summaries that only include source simplification and elaborative explanation sentences are 198 and 191 respectively, and 198 and 194 for non-factual summaries. Simplification: Source Simplification; Explanation: Elaborative Explanation; $\tau$: Kendall's~$\tau$; \textsc{Abs}: abstracts; \textsc{TB}: Textbooks; \textsc{SP}: StatPearls; AUC: AUC-ROC. * indicates a statistically significant difference compared to \textsc{Abs+TB+SP} ($p<0.01$). We run each setting for five times and report the standard deviation (std.) in the brackets.}
\label{tab:retrieval-ablation}
\end{table}

\noindent \textbf{Combined Domain Resources for Retrieval}
\rev{As shown in Table~\ref{ablation_study}, combining all three evidence sources (\textsc{Abs+TB+SP}) significantly outperforms the other two settings ($p<0.01$). This suggests that comprehensive retrieval, which includes the original source abstracts, improves overall factual consistency assessment. Table~\ref{tab:retrieval-ablation} shows the retrieval ablation study results in detail. In the full dataset setting, \textsc{Abs+TB+SP} achieves AUC-ROC=$96.4$ and Pearson=$81.3$, outperforming all retrieval variants that use only a subset of the retrieval corpus, including \textsc{TB+SP} (AUC-ROC=$94.8$), \textsc{TB} only ($94.0$), \textsc{SP} only ($93.3$), and \textsc{Abs+TB} ($95.4$).}  To further examine the effectiveness of retrieval for source simplification and elaborative explanation, Table~\ref{tab:retrieval-ablation} breaks down the results as follows:

(1) \emph{Source simplification}: These are sentences in plain language summaries classified as ``source simplification.'' We evaluate summaries containing only simplification sentences, using abstracts as the source. Using abstracts as sole source texts achieves strong performance (63.5/79.0/94.8) across all three criteria.

(2) \emph{Elaborative explanation}: For summaries containing only explanation sentences, abstracts alone are not sufficient, performing much worse than in simplification cases. \rev{Adding abstracts with external sources improves performance, with the best results obtained by combining all three sources. \textsc{Abs+TB+SP} outperforms the abstract-only baseline \textsc{Abs} (AUC-ROC=$72.7$) and all partial-view variants (\textsc{TB+SP}, \textsc{TB}, \textsc{SP}, \textsc{Abs+TB}, and \textsc{Abs+SP}). The gain over \textsc{Abs} is substantial, with $\Delta\mathrm{AUC-ROC} = 8.3$ on (99\% CI [3.7, 13.0]), and similarly large improvements in correlation.}

(3)\emph{Full dataset}: Retrieval from StatPearls provides better results than retrieval from Textbooks, highlighting the importance of using high-quality, domain-specific knowledge bases for PLS evaluation. Overall, the best combination across all settings is \textbf{\textsc{Abs+TB+SP}} (65.7/81.3/96.4). \rev{Relative to the abstract-only baseline, \textsc{Abs} (AUC-ROC=$94.3$), the gain in discrimination is $\Delta\mathrm{AUC-ROC} = 2.1$ (99\% CI [0.5, 4.2], where $\Delta$ is \textsc{Abs+TB+SP} minus \textsc{Abs}).}

These findings show that effective retrieval for factual consistency should include both the source abstract and reliable external medical references, especially when summaries contain elaborative explanations.

\noindent \textbf{Answer Overlap Evaluation}
\rev{BERTScore ties with the LERC approach introduced in QAFactEval~\cite{fabbri2022qafacteval}, showing no significant difference in AUC-ROC ($p=0.29$). BERTScore achieves slightly higher nominal values in both correlation metrics. There are fundamental differences between the two metrics.} LERC is a learned model trained to predict human factual consistency scores, which may reduce its generalizability to new domains or sentence types. In contrast, BERTScore computes embedding-based semantic similarity, allowing it to capture fine-grained semantic overlap between generated and gold answers. Since \factscore evaluates factual consistency by comparing model-generated answers to those extracted from the source, BERTScore’s sensitivity to semantic alignment makes it a more effective and robust choice for this step of the evaluation.

\noindent \textbf{Sentence-level vs.\ Summary-level Evaluation}
\rev{To further examine how input granularity affects factual consistency evaluation, we deactivate the sentence-splitting function and instead pair each plain language summary with its abstract to evaluate factual consistency at the summary level. Our best sentence level evaluation achieves an AUC-ROC of $96.4$, which is significantly higher than the summary level setting ($p<0.01$). The summary level setting performs worse across all three criteria (Kendall's~$\tau$=$49.3$, Pearson=$59.6$, AUC-ROC=$84.7$; std.\ $0.5$). }Evaluating at the summary level introduces broader contextual dependencies, which can omit sentence-specific factual errors or create inaccurate entailments. \rev{These results suggest that processing one sentence at a time makes QA-based factual consistency evaluation more focused and reduces noise from unrelated context.}

\begin{table}[!ht]
\scriptsize 
\setlength{\tabcolsep}{6pt} 
\renewcommand{\arraystretch}{1.1} 
\resizebox{\textwidth}{!}{ 
    \begin{tabularx}{\textwidth}{@{}lX@{}}
    \toprule
    \multirow{9}{*}{\rotatebox{90}{\textbf{Correct Case}}} & \cellcolor[HTML]{D9E1F2}\textbf{Plain Language Sentence:} Limits to the availability of \ctext{245,220,120}{SSB} in schools (e.g. replacing SSBs with water in school cafeterias). \cite{Von-Philipsborn19} \\
     & \textbf{Scientific Abstract:} Frequent consumption of excess amounts of sugar‐sweetened beverages (SSB) is a risk factor for obesity, type 2 diabetes, cardiovascular disease and dental caries... \cite{Von-Philipsborn19} \\
     & \textbf{Generated Question:} Limits to the availability of what in schools?\\
     & \textbf{Retrieved Knowledge:} Even though additional data is required to determine the impact of limiting the availability of \ctext{160, 205, 205}{nutrient-poor or high-sugar goods} in schools on obesity, some study results have shown a net-positive result...\\
     & \textbf{Final Answer:} \ctext{180, 220, 100}{nutrient-poor or high-sugar goods}\\ 
     \cmidrule(l){2-2}
    \multirow{8}{*}{\rotatebox{90}{\textbf{Misclassification}}} & \cellcolor[HTML]{D9E1F2}\textbf{Plain Language Sentence:} This review looked at how well the methods worked to prevent \ctext{245,220,120}{pregnancy}, if they caused bleeding problems, if women used them as prescribed, and how safe they were. \cite{Lopez03} \\
     & \textbf{Scientific Abstract:} To compare the contraceptive effectiveness, \ctext{160, 205, 205}{cycle control}, compliance (adherence), and safety of the contraceptive patch or the vaginal ring versus combination oral contraceptives (COCs)... \cite{Lopez03} \\
     & \textbf{Generated Question:} This review looked at how well the methods worked to prevent what?\\
     & \textbf{Retrieved Knowledge:} Appropriate treatment for the underlying etiology should start as soon as possible, and the patients and family members should receive appropriately targeted education...\\
     & \textbf{Final Answer:} \ctext{255, 190, 180}{cycle control}\\ 
     \cmidrule(l){2-2}
     \multirow{10}{*}{\rotatebox{90}{\textbf{Retrieval Quality}}} & \cellcolor[HTML]{D9E1F2}\textbf{Plain Language Sentence:} The patch is a small, thin, \ctext{245,220,120}{adhesive} square that is applied to the skin. \cite{Lopez03} \\
     & \textbf{Scientific Abstract:} Users of the norelgestromin‐containing patch reported more breast discomfort, dysmenorrhea, nausea, and vomiting. In the levonorgestrel‐containing patch trial, patch users reported less vomiting, headaches, and fatigue...\cite{Lopez03} \\
     & \textbf{Generated Question:} The patch is a small, thin, what kind of square applied to the skin?\\
     & \textbf{Retrieved Knowledge:} \ctext{160, 205, 205}{Nonstick dressing Petrolatum-infused gauze strip} or other material to form a bolster over the graft site. This may be sutured or taped securely in place to provide some pressure and to keep graft immobilized.\\
     & \textbf{Final Answer:} \ctext{255, 190, 180}{Nonstick dressing Petrolatum-infused gauze strip}\\ 
     \cmidrule(l){2-2}
     \multirow{13}{*}{\rotatebox{90}{\textbf{Unanswerable Question}}} & \cellcolor[HTML]{D9E1F2}\textbf{Plain Language Sentence:} \ctext{245,220,120}{Government} officials, business people and health professionals implementing such measures should work together with researchers to find out more about their effects in the short and long term. \cite{Von-Philipsborn19} \\
     & \textbf{Scientific Abstract:} To assess the effects of environmental interventions (excluding taxation) on the consumption of sugar‐sweetened beverages and sugar‐sweetened milk, diet‐related anthropometric measures and health outcomes, and on any reported unintended consequences or adverse outcomes...\cite{Von-Philipsborn19}\\
     & \textbf{Generated Question:} What officials, business people and health professionals implementing such measures should work together with researchers to find out more about their effects in the short and long term? \\
     & \textbf{Retrieved Knowledge:} Implementation should be accompanied by high‐quality evaluations using appropriate study designs, with a particular focus on the \ctext{160, 205, 205}{long‐term effects} of approaches suitable for large‐scale implementation.\\
     & \textbf{Final Answer:} \ctext{255, 190, 180}{long‐term effects}\\
    \bottomrule
    \end{tabularx}
} % end resizebox
\caption{Error analysis of \factscore with intermediate metric outputs. Retrieved Knowledge refers to the source texts retrieved for each plain language sentence. We present correct and failure cases sampled only from elaborative explanation examples (i.e., sentences classified as explanations during evaluation). The correct case has a \factscore score above 0.6, while the failure cases have scores below 0.5. 
\newline \footnotesize \textbf{Color legend:} 
\footnotesize \ctext{245,220,120}{extracted answer}, 
\footnotesize \ctext{160,205,205}{answer origin},
\footnotesize \ctext{180,220,100}{correct answer}, 
\footnotesize \ctext{255,190,180}{incorrect answer}.
}
\label{tab:error-analysis}
\end{table}

\subsection{Error Analysis on \benchmark} \label{sec:error_analysis}
To analyze cases where \factscore fails on the \benchmark benchmark, we categorize one correct example and three types of errors in Table~\ref{tab:error-analysis}. For each case, we present the original plain language sentence, the corresponding scientific abstract with retrieved knowledge (from medical Textbooks and StatPearls), the model-generated questions based on extracted answers, and the QA model’s final responses. Each question is generated based on the extracted answer and its corresponding plain language sentence. 

In the correct case, the QG model generates a question from the plain language sentence and the extracted answer ``SSB.'' The QA model then provides the correct answer using the retrieved knowledge rather than the original abstract, demonstrating \factscore's effectiveness in evaluating explanation sentences.

In the first failure case, the extracted answer is ``pregnancy,'' but the QA model returns ``cycle control'' from the abstract because the correct term is missing in the retrieved content. Since this sentence is classified as an ``elaborative explanation,'' which requires external knowledge for verification, this error points to a potential misclassification between \textit{simplification} and \textit{explanation} by the fine-tuned classifier. The second failure illustrates how noisy retrieved knowledge can impair evaluation. The QA model provides an irrelevant answer from the retrieved content that does not match the extracted gold answer. The third case shows an unanswerable question generated by the QG model (e.g., regarding ``Government''), highlighting a limitation of QA-based factual consistency evaluation. Some questions remain unanswered even with retrieval. Additionally, we acknowledge that some plain language sentences may not generate any questions, resulting in empty question sets. Addressing these challenges requires further improvements in classification, retrieval, and domain-specific question generation.

\subsection{\rev{Error Analysis on FareBio}} \label{error-analysis-farebio}
\begin{table}[!ht]
\centering
\scriptsize % Slightly larger than scriptsize for readability
\setlength{\tabcolsep}{5pt}
\renewcommand{\arraystretch}{1.3} % More breathing room between rows
\newcommand{\qacell}[3]{%
    \textbf{Ref:} #1 \newline
    \textbf{Q:} \textit{#2} \newline
    \textbf{Pred:} #3%
}
% \color{orange}
\begin{tabularx}{\textwidth}{@{} l >{\RaggedRight\arraybackslash}p{3.8cm} >{\RaggedRight\arraybackslash}X >{\RaggedRight\arraybackslash}X @{}}
\toprule
\textbf{Type} & \textbf{Target PLS Sentence} & \textbf{QA Extraction Analysis} & \textbf{Rationale} \\
\midrule

% Row 1
\textbf{NF} & 
This performance was identical in the comparative analysis subgroup and was within the range of physicians at different levels of expertise: 0.86-0.97 and 0. & 
\qacell{physicians}
{This performance was identical in the comparative analysis subgroup and was within the range of what at different levels of expertise?}
{physicians} & 
The non-factual error is a value mistake. The target sentence incorrectly truncates the specificity range to ``0,'' which does not appear in the source and therefore being labeled as non-factual. However, our answer extractor fails to identify the numerical values as gold answer to verify its factual consistency.\\
\addlinespace[1em] % Extra space between distinct entries

% Row 2
\textbf{NF} & 
Imaging also showed chronic compression of the coeliac axis with compensatory hypertrophy of the gastroduodenal artery. & 
\qacell{gastroduodenal artery}
{Imaging also showed chronic compression of the coeliac axis with compensatory hypertrophy of what artery?}
{gastroduodenal} & 
The target sentence correctly restates information that is directly present in the source, but it is labeled as non-factual in FareBio. \\
\addlinespace[1em]

% Row 3
\textbf{F} & 
In summary, the article discusses the potential impact of COVID-19 on the musculoskeletal system... and emphasizes the need for further research to understand the underlying mechanisms and develop appropriate interventions. & 
\qacell{interventions}
{In summary, the article discusses the potential impact of COVID-19 on the musculoskeletal system, particularly in the context of aging and Long-COVID, and emphasizes the need for further research to understand the underlying mechanisms and develop appropriate what?}
{multidisciplinary approach} & 
The predicted answer comes from the original source. This target PLS sentence is classified as a simplification; however, the term ``interventions'' does not appear in the source.\\
\addlinespace[1em]

% Row 4
\textbf{F} & 
Some patients experience persistent symptoms after recovering from the initial infection, a condition known as ``Long-COVID.'' & 
\qacell{infection}
{Some patients experience persistent symptoms after recovering from what condition known as Long-COVID?}
{SARS-CoV-2} & 
In the source, SARS CoV 2 always appears together with the word ``infection.'' Our QA model captures the relevant answer, but our answer overlap evaluation approach does not align this prediction with the gold answer, which results in a lower \factscore.\\

\bottomrule
\end{tabularx}
\caption{Error analysis of \factscore on the FareBio dataset. NF/F = non-factual or factual; Ref = gold/extracted answer; Q = generated question; Pred = predicted answer.}
\label{tab:farebio-error-analysis}
\end{table}

\rev{As shown in Table~\ref{tab:farebio-error-analysis}, we provide several failure cases on the FareBio dataset based on the results in Table~\ref{tab:main-results}. We rank the \factscore scores for all summaries and select two examples with the lowest scores among those labeled NF and two examples with the lowest scores among those labeled F. The first NF example in Table~\ref{tab:farebio-error-analysis} shows a true factual error. The target plain language sentence changes the numerical range and introduces the value zero, which does not appear in the source. However, our answer extractor does not reliably extract the exact numbers, so \factscore does not capture the fact errors. The second NF example shows a label noise in FareBio. The target sentence restates information that is clearly present in the source, but the sentence is labeled as NF. Third, we find pipeline errors even when the FareBio label is factual. In the third example, the target sentence correctly summarizes the current knowledge and is classified as simplification by our classifier, but the answer extractor extracts the word ``interventions,'' which does not appear in the source and thus leads to a low score. In the fourth example, the QA module predicts ``SARS CoV 2'' as the answer, which is consistent with the ``infection'' described in the plain language sentence, but the final answer overlap evaluation step cannot match this prediction to the gold answer ``infection.'' Overall, we conclude that most errors of our \factscore come from the answer extraction and answer overlap evaluation stages in the FareBio dataset, while the dataset contains noisy factuality labels.}

\rev{To mitigate these errors (\S\ref{sec:error_analysis}, \S\ref{error-analysis-farebio}), we provide several directions for future work. First, misclassification of plain language sentences by the factuality type classifier can be reduced by training on more annotated examples or use advanced low-resource model tuning strategies for text classification \cite{you-etal-2024-sciprompt}. Second, retrieval errors can be mitigated by conducting precise filtering method to select the most relevant snippets, such as limiting the length of retrieved snippets and chunk size, using a combination of dense and sparse retrieval, and expanding the domain corpus (e.g., PubMed) to cover more specialized topics. Third, to handle unanswerable or vague questions, future work can improve the question filtering stage that removes questions that do not have clear grounding in both source retrieved context, or that are too general to support reliable factual checks using heuristic approaches. Finally, for the answer overlap evaluation module, more advanced entailment-based models can be applied to replace the BERTScore semantic matching to check whether the predicted answer is consistent with the gold answer after simple normalization of entities and synonyms. \orange{We report the quantitative error analysis of FareBio in ~\ref{app:quantative-error-analysis}.}}

\section{Discussion and Conclusion}
Our study advances the assessment of hallucinations in the field of PLS by supplying both \purple{a} domain expert-annotated biomedical PLS benchmark \benchmark and a novel retrieval‐augmented QA‐based factual consistency evaluation metric \factscore. Unlike existing biomedical domain corpora that either lack fine‐grained labels for elaborative content \cite{luo2024factual} or focus solely on text simplification (e.g., FactPICO\cite{joseph-etal-2024-factpico}), \benchmark captures sentence‐level distinctions, including simplification versus explanation, functional roles, and explicit alignment to source sentences. This granularity not only enables precise error analysis of hallucinations introduced for clarity, but also provides a reusable framework for researchers in other disciplines (e.g., legal, technical) to replicate our annotation protocol and build high‐quality factual consistency datasets tailored to their specialized texts. By offering these data and detailed annotation protocol (\ref{appx:annotation-protocol}), we anticipate community‐driven extensions, such as multilingual adaptations or integration with domain‐specific ontologies that will democratize development of trustworthy summarization models.

Building on this resource, \factscore differentiates from prior QA‐based metrics (e.g., QAFactEval\cite{fabbri2022qafacteval}, QuestEval\cite{scialom-etal-2021-questeval}) by first classifying whether a sentence requires external verification given elaborative explanations are often ignored by metrics relying solely on source text, and improve the evaluation \orange{accuracy} without retrieving for every plain language summary instance. Then, retrieving targeted domain knowledge for each elaborative explanation before posing and answering fact‐checking questions. The ablations demonstrate that selective retrieval yields substantial gains in Kendall's~$\tau$, Pearson correlation, and AUC‐ROC over both end‑to‑end LLM judges\cite{llama3-2024} and alignment‐based scorers\cite{zha2023alignscore}, highlighting that the incorporation of external evidence is essential for robust factual consistency assessment. Beyond its empirical strengths, \factscore serves as a plug‑and‑play evaluation metric for future summarization systems. Developers can leverage our classifier to classify sentences, apply retrieval only where necessary, thus containing computational costs and obtain interpretable question–answer pairs that manifest non-factual hallucinations. More importantly, \factscore is developed with the open-source backbone model (i.e., \texttt{Llama 3.1}), ensuring broad use with transparency, reproducibility, and lower computational costs. \purple{Looking forward, our proposed evaluation metric could serve as a diagnostic tool for researchers and developers to identify and analyze specific factual inconsistencies in PLS systems, guiding future improvements in both model development and evaluation methodology.}

\purple{Despite these promising results, we consider the following limitations of this study: (1) Our error analysis reveals that misclassifications, particularly between source simplifications and elaborative explanations, can lead to retrieval mismatches and unanswerable QA prompts. These issues underscore the need for more nuanced classification and question generation modules. Given the fair IRA scores, especially for the factuality type field, we acknowledge that our classification method is not fully reliable. The annotated dataset reflects subjective judgments, particularly when distinguishing source simplification from elaborative explanation, and the training set is limited in size. We advise users to apply our pretrained classifier with caution when using it for later factual consistency evaluation steps. Future efforts are needed on developing domain-specific (e.g., law, healthcare, social science, etc.) classifiers \cite{you-etal-2024-sciprompt} for factual consistency evaluation. (2) While our experiments indicate that our retrieval-augmented evaluation metric can improve factual consistency assessment in most of the settings, especially in elaborative explanation evaluation, the computational time increases compared to NLI-based evaluation metrics and LLM-based evaluators. The trade-off between evaluation precision and efficiency suggests that further optimization will be beneficial. We suggest to balance wisely based on the trade-offs we report in this study regarding the evaluation time and accuracy. \orange{\factscore is a research-oriented metric that prioritizes interpretability and transparency over speed.} \rev{(3) The quantitative error analysis on FareBio (Table~\ref{tab:farebio-quantitative-error}) shows a notable limitation of \factscore in handling numerical content. The answer extraction step may overlook numerical values, preventing them from being verified in subsequent question generation and answering stages. Consequently, \factscore may underestimate factual inconsistency in summaries with incorrect numerical information.} \purple{(4) \benchmark and \factscore are built upon the CELLS corpus, whose plain language summaries are written by the original article authors. While this ensures domain expertise, it does not guarantee complete factual consistency, as expert-authored summaries may contain unintentional inaccuracies or oversimplifications.}}

% Although we provide substitute solutions for classifying input summaries or sentences, our fine-tuned classifier is limited to the biomedical domain and specifically designed for PLS tasks, fine-tuned through \benchmark.

\section*{Acknowledgment}
This work used Delta GPU at NCSA through allocation [CIS240504] from the Advanced Cyberinfrastructure Coordination Ecosystem: Services \& Support (ACCESS) program, which is supported by U.S. National Science Foundation grants \#2138259, \#2138286, \#2138307, \#2137603, and \#2138296.

\clearpage
\appendix
\section{Dataset Annotation Protocol}
\label{appx:annotation-protocol}
We develop a comprehensive annotation procedure for freelancers on Upwork to conduct fact-checking annotations. 

Our annotation procedure involves two stages, starting with thorough training using detailed examples to ensure consistent understanding of the task between annotators. 
Annotators receive a spreadsheet where each row contains a pair of data: a sentence extracted from the plain language summary and its corresponding scientific abstract. For every sentence-abstract pair, the annotators are required to label three features: Factuality type, Functional role, and Sentence alignment, with the appropriate labels.

\begin{tcolorbox}[
  enhanced,
  breakable,
  boxrule=1pt,
  % colback=cadmiumgreen!5!white,
  % colframe=forestgreen!75!white,
  boxsep=5pt,
  arc=4pt,
  title=\textit{Step 1: Sentence Annotation Instructions}
]
\footnotesize
Compared to the scientific abstract, analyze each sentence of the plain language summary across three dimensions: external information, category, and relation.\\

\noindent 1. \textbf{Factuality type}: Determine whether the sentence includes information does not present in the scientific abstract.
\begin{enumerate}
\item[1)] \texttt{Yes}: The sentence contains external information that is not explicitly mentioned, paraphrased, or implied in the scientific abstract.
\item[2)] \texttt{No}: The sentence contains information that is explicitly stated or closely paraphrased from the scientific abstract.
\end{enumerate}

\noindent 2. \textbf{Functional role}: Classify the sentence of the plain language summary into one of the following categories (you can only choose one category per annotation)
\begin{enumerate}
\item[1)] \texttt{Definition}: Provides a fundamental explanation of a term.
\item[2)] \texttt{Background}: Information that helps understand the term within the context of the abstract, such as relevance, significance, or motivation.
\item[3)] \texttt{Example}: Specific instances that illustrate the use of the term in the scientific abstract.
\item[4)] \texttt{Method/result}: Details about the methodology or results described in the scientific abstract.
\item[5)] \texttt{Other}: For sentences that do not fit into the categories above, please indicate the category
\end{enumerate}

\noindent 3. \textbf{Sentence alignment}: Identify the sentence(s) in the scientific abstract that the sentence of the plain language summary is related to. Use indices of sentences from the scientific abstract to link the sentence of the plain language summary. You can select one or more sentences from the scientific abstract. List the indices like s1\_1,s2\_3,s3\_6. If no relation is found, mark it as ``external.''
\end{tcolorbox}

In Step 2, we provide explanations of the existing sentence-level indexes for plain language sentences and scientific abstracts. 

\begin{tcolorbox}[
  enhanced,
  breakable,
  boxrule=1pt,
  boxsep=5pt,
  arc=4pt,
  title=\textit{Step 2: Using the Annotation Spreadsheet}
]
\footnotesize
You will work within a structured spreadsheet containing the segmented sentences from both the summaries and the scientific abstracts. \\

\noindent 1. \textbf{Target\_Summary\_ID}: A unique identifier for each plain language summary. Sentences from the same plain language summary share the same ID.\\

\noindent 2. \textbf{Target\_Sentence\_Index}: An identifier for each sentence within a summary, forming as tx\_y, where `tx' is the same as its `Target\_Summary\_ID', and `y' represents the index of each sentence in the plain language summary, starting from 1. e.g., t0\_1 refers to the first sentence of the first summary)\\

\noindent 3. \textbf{Target\_Sentence}: The plain language sentence you are annotating.\\

\noindent 4. \textbf{Original\_Abstract}: The abstract corresponding to each summary, with each sentence indexed for easy reference.

\end{tcolorbox}

Each annotator will annotate 40 summary-abstract pairs to ensure each sentence of the plain language summary has two sets of annotations from different people. For each row of the spreadsheet, they need to annotate three columns: ``Factuality type,'' ``Functional role,'' and ``Sentence alignment.''

\begin{tcolorbox}[
  enhanced,
  breakable,
  boxrule=1pt,
  boxsep=5pt,
  arc=4pt,
  title=\textit{Annotation Fields to Be Completed}
]
\footnotesize
\noindent \textcolor{red}{$<$TO BE ANNOTATED$>$} \textbf{Factuality type}: Mark ``Yes'' or ``No'' to indicate if the sentence in the `Target\_Sentence' column contains external information. \\

\noindent \textcolor{red}{$<$TO BE ANNOTATED$>$} \textbf{Functional role}: Choose the most fitting category of the sentence from the list (Definition, Background, Example, Method/Result, Other).\\

\noindent \textcolor{red}{$<$TO BE ANNOTATED$>$} \textbf{Sentence alignment}: List the relevant sentence indices from the abstract in the `Original\_Abstract' column that relate to the plain language sentence (e.g., s10\_1,s10\_5). For example, filling in `s10\_1,s10\_5,s10\_9' if you think these three sentences from the abstract are relevant to sentence t1\_3. Use commas to separate multiple indices. 
\end{tcolorbox}

To ensure annotators fully understand the context and task requirements, we provide comprehensive annotation training and a screening test prior to the annotation process. We select candidates from freelancers with a medical education background, and only those who pass the screening test are finalized as annotators.

To ensure annotators fully understand the context and task requirements, we provide comprehensive annotation training and a screening test prior to the annotation process. We select candidates from freelancers with a medical education background, and only those who pass the screening test are finalized as annotators.

\section{Dataset Examples} \label{appx:dataset-examples}
\begin{table}[!ht]
\scriptsize % smaller than \small
\setlength{\tabcolsep}{3pt} % reduce column padding
\renewcommand{\arraystretch}{0.9} % tighten row spacing

\resizebox{\textwidth}{!}{ % ensure it fits the page
    \begin{tabularx}{\textwidth}{@{\extracolsep\fill}ll|X}
    \toprule
     & \textbf{Feature} & \textbf{Annotation} \\
    \midrule
    \multirow{9}{*}{\rotatebox{90}{\textbf{Example 1}}} 
     & Plain Language Sentence & Gout caused by crystal formation in the joints due to high uric acid levels in the blood.\\
    \cmidrule(ll){2-3}
     & Need External Information? & yes \\
    \cmidrule(ll){2-3}
     & Functional role & Background \\
    \cmidrule(ll){2-3}
     & Sentence alignment & external \\
    \cmidrule(ll){2-3}
     & Corresponding Abstract & None\\
    \midrule \\[-0.8em] % tighten spacing between examples
    \midrule
    \multirow{11}{*}{\rotatebox{90}{\textbf{Example 2}}} 
     & Plain Language Sentence & Reducing blood pressure with drugs has been a strategy used in patients suffering from an acute event in the heart or in the brain, such as heart attack or stroke.\\
    \cmidrule(ll){2-3}
     & Need External Information? & yes \\
    \cmidrule(ll){2-3}
     & Functional role & Background \\
    \cmidrule(ll){2-3}
     & Sentence alignment & s10\_1,s10\_2 \\
    \cmidrule(ll){2-3}
     & Corresponding Abstract & $<$s10\_1$>$Acute cardiovascular events represent a therapeutic challenge. $<$s10\_2$>$Blood pressure lowering drugs are commonly used and recommended in the early phase of these settings.  \\
    \midrule \\[-0.8em]
    \midrule
    \multirow{18}{*}{\rotatebox{90}{\textbf{Example 3}}} 
     & Plain Language Sentence & We looked at whether choice of antibiotic made a difference in the number of people who experienced failed treatment, and we determined the proportions who had resolution of fever at 48 hours. \\
    \cmidrule(ll){2-3}
     & Need External Information? & no \\
    \cmidrule(ll){2-3}
     & Functional role & Method/Result \\
    \cmidrule(ll){2-3}
     & Sentence alignment & s15\_16,s15\_17,s15\_20 \\
    \cmidrule(ll){2-3}
     & Corresponding Abstract & $<$s15\_16$>$For treatment failure, the difference between doxycycline and tetracycline is uncertain (very low‐certainty evidence). $<$s15\_17$>$Doxycycline compared to tetracycline may make little or no difference in resolution of fever within 48 hours (risk ratio (RR) 1.14, 95\% confidence interval (CI) 0.90 to 1.44, 55 participants; one trial; low‐certainty evidence) and in time to defervescence (116 participants; one trial; low‐certainty evidence). $<$s15\_20$>$For most outcomes, including treatment failure, resolution of fever within 48 hours, time to defervescence, and serious adverse events, we are uncertain whether study results show a difference between doxycycline and macrolides (very low‐certainty evidence). \\
    \bottomrule
    \end{tabularx}
} % end resizebox
\caption{Examples of our curated dataset. Need External Information feature represents whether a plain language sentence is a simplification or an explanation. A label of ``yes'' indicates that the sentence is an explanation and requires additional elaborative information beyond the source abstract to verify its factual consistency. Conversely, a label of ``no'' shows that the sentence is a simplification that can be validated using only the source abstract.}
\label{tab:data-examples}
\end{table}

In Table~\ref{tab:data-examples}, we presents three representative examples from \benchmark. Each example is annotated with five features: a plain language sentence, an indicator of whether the sentence is simplification or explanation, its category, its relation, and the corresponding abstract. All plain language sentences are \textbf{factual}, as they were written by the authors from the Cochrane database. ``Need External Information?'' feature specifies whether a sentence can be validated solely by the abstract. A ``Yes'' label indicates that the sentence includes information not explicitly mentioned in the abstract, and vice versa. The ``Sentence alignment'' feature identifies the sentence(s) in the scientific abstract most relevant to the plain language summaries; if no corresponding content exists in the abstract, the relation is marked as ``external.'' We randomly sample three examples from the dataset to to illustrate the dataset's structure. Additionally, indexes have been created for both the plain language sentences and the abstract sentences to facilitate annotation.

\section{\rev{Qualitative Analysis of \benchmark}} \label{appx:IAA-qualitative}
\begin{table}[!ht]
\centering
\scriptsize
\setlength{\tabcolsep}{4pt}
\renewcommand{\arraystretch}{1.2}
% \color{orange} %%%%% remove this
\begin{tabularx}{\textwidth}{@{}l>{\raggedright\arraybackslash}p{4.2cm}ccc>{\raggedright\arraybackslash}X@{}}
\toprule
\textbf{ID} & \textbf{Target PLS Sentence} & \textbf{A1} & \textbf{A2} & \textbf{Final} & \textbf{Rationale} \\
\midrule
\textbf{Ex1} 
 & Obesity is associated with many health problems and a higher risk of death.
 & \xmark & \cmark & \cmark 
 & ``Health problems'' and ``higher risk of death'' are not in the original abstract. No association between obesity and health/death issues exists in the source. \\
\addlinespace[0.5em]
\textbf{Ex2} 
 & It can be hard for the person to gain control of the eye, which can cause distress in social situations.
 & \cmark & \xmark & \cmark 
 & Most relevant source (from A1): \textit{``Spontaneously manifest DVD can be difficult to control and often causes psychosocial concerns.''} However, no direct correlation between eye control and DVD exists; social situations are not equal to psychosocial concerns.\\
\addlinespace[0.5em]
\textbf{Ex3} 
 & Most participants were middle-aged.
 & \cmark & \xmark & \xmark 
 & A2 provides supporting source: \textit{``Applicability of evidence was limited to middle-aged participants who were relatively free of co-morbidity and were undergoing elective abdominal surgery.''} This sentence is supported by the source.\\
\addlinespace[0.5em]
\textbf{Ex4} 
 & In some people, winter blues becomes depression, which seriously affects their daily lives.
 & \cmark & \xmark & \cmark 
 & A2 highlights relevant source: \textit{``SAD is a seasonal pattern of recurrent major depressive episodes that most commonly occurs during autumn or winter and remits in spring.''} However, no causal relation between winter blues and depression in source.\\
\addlinespace[0.5em]
\textbf{Ex5} 
 & RDS is the most common cause of disease and death in babies born before 34 weeks' gestation.
 & \xmark & \cmark & \cmark 
 & Most relevant source: \textit{``RDS is the single most important cause of morbidity and mortality in preterm infants.''} However, ``babies born before 34 weeks' gestation'' is not mentioned in the abstract. \\
\bottomrule
\end{tabularx}
\caption{Qualitative analysis of \benchmark annotation examples. $A1/A2$ = Annotator 1 or 2; \cmark/\xmark \space indicate yes or no judgments for the ``Need External Information?'' field.}
\label{tab:IAA-qualitative}
\end{table}

\rev{Given the fair inter-rater agreement for the “Need External Information?” field in \benchmark, we conduct a targeted qualitative error analysis. We randomly sample one representative plain language sentence from each of five annotation batches where the two annotators disagree. Two annotators independently judge whether external information is required, and a third judge provides a brief rationale. As shown in Table~\ref{tab:IAA-qualitative}, most disagreements fall into two categories: (i) differing interpretations of whether an elaborative explanation introduces new information beyond the source, and (ii) failures to locate or recognize supporting evidence in the source abstract. These examples highlight that the “Need External Information?” decision often depends on subtle judgments about sufficiency and specificity of source coverage, suggesting that clearer guidelines and additional training examples may further improve annotation consistency.}

\section{\rev{Data and Indexing Statement}} \label{appx:retrieval-statement}
\rev{In this study, we follow the same protocol as used in MedRAG \cite{xiong2024benchmarking} to conduct sentence-level retrieval of the PLS sentences. Specifically, we use top-K as 3 across all of our experiments on both StatPearls\footnote{\url{https://huggingface.co/datasets/MedRAG/statpearls}} and medical Textbooks\footnote{\url{https://huggingface.co/datasets/MedRAG/textbooks}} \cite{jin2021disease} provided in the MedRAG repository\footnote{\url{https://github.com/Teddy-XiongGZ/MedRAG?tab=readme-ov-file\#corpus}}. Specifically, the medical Textbooks corpus comprises 18 widely used medical textbooks (e.g., for USMLE preparation) originally collected by \citet{jin2021disease}, while StatPearls corpus consists of point-of-care clinical decision support articles. The authors of MedRAG collect 9,330 publicly available articles from the NCBI Bookshelf\footnote{\url{https://www.ncbi.nlm.nih.gov/books/NBK430685/}}, \orange{which permits redistribution for non-commercial research purposes.} The StatPearls corpus applies a hierarchical approach for chunking, treating each paragraph in an article as a snippet and splice all relevant hierarchical headings to form the corresponding title. The medical Textbooks is processed into chunks of no more than 1,000 characters. \orange{These corpus are made publicly available for research replication via the MedRAG HuggingFace repository\footnote{\url{https://huggingface.co/MedRAG}} under the original authors' distribution terms. We do not redistribute raw StatPearls and Textbooks content beyond the MedRAG release and rely exclusively on their officially published corpus to ensure replicability. Researchers wishing to replicate the corpus index should (i) access the same StatPearls\footnote{\url{https://huggingface.co/datasets/MedRAG/statpearls}} and Textbooks\footnote{\url{https://huggingface.co/datasets/MedRAG/textbooks}} datasets, (ii) ensure their use complies with the copyright and permissions of the source license of MedRAG \cite{xiong2024benchmarking} and MedQA \cite{jin2021disease}.} Following MedRAG, we use the same \texttt{RecursiveCharacterTextSplitter} from the LangChain\footnote{\url{https://www.langchain.com/}} library to perform chunking. We directly retrieve top 3 most relevant snippets from both corpus. \orange{For the retriever, we use MedCPT \cite{jin2023medcpt} (specifically the \texttt{ncbi/MedCPT-Query-Encoder\footnote{\url{https://huggingface.co/ncbi/MedCPT-Query-Encoder}}} version, commit hash: \texttt{d83a36c}) for dense retrieval with CLS token pooling. Documents are indexed using FAISS \cite{johnson2021billion} with an \texttt{IndexFlatIP} configuration (exact inner product search, , no quantization or clustering) for 768-dimensional embeddings. The retrieval system computes the inner product between query and document vectors to identify the most relevant snippets. This matches the default MedRAG implementation for MedCPT.}}

\section{\rev{Reproducibility}} \label{appx:experiment-settings}

\subsection{\rev{Fine-tuning QG Model}}
\rev{We fine-tune the QG model (i.e., \texttt{BART-large}) with the same datasets used in QAFactEval \cite{fabbri-etal-2022-qafacteval}, including SQuAD \cite{rajpurkar-etal-2016-squad} and QA2D \cite{demszky2018transforming}. We use a batch size of 16, a learning rate of 3e-5, maximum sequence length of 512 tokens, and two training epochs for QG model fine-tuning.}

\subsection{\rev{Fine-tuning Factuality Type Classifier}}
\rev{As mentioned in Section 3.2.1, we fine-tune a classifier using \texttt{PubMedBERT-base} model through our curated \benchmark dataset. Specifically, we set the random seed to 42 to split \benchmark into training, validation, and test sets.  We tune the \texttt{PubMedBERT-base} model for three epochs with early stopping under the validation set. The final fine-tuning accuracy of the classifier on the test set is 0.77. The same seed (42) is also used to sample 200 summary-abstract pairs from the CELLS test set for the comparison in Table~\ref{tab:main-results}. The classifier used in \factscore is fine-tuned on \benchmark with a batch size of 32 for 10 epochs. We apply early stopping during training based on validation loss.}

\subsection{\rev{General Details}}
\rev{For other open-source models used in this study, we apply the default version. For the QA model, we apply the original \texttt{Electra-large} model\footnote{Can be downloaded here: \url{https://github.com/salesforce/QAFactEval/blob/master/download_models.sh}} used in QAFactEval \cite{fabbri-etal-2022-qafacteval}. All experiments we conduct are under one NVIDIA A100 GPU with 40 GB GPU memory. We employ the Natural Language Toolkit (NLTK)\footnote{\url{https://www.nltk.org/}} to split the plain language summaries into sentences using the \texttt{punkt} package. For question filtering strategy, we follow the same protocol in QAFactEval. The temperature is set to 0 for \texttt{GPT-4o} and 0.01 for \texttt{Llama 3.1}. The \texttt{Llama 3.1} model used in AE also uses a temperature of 0.01. The maximum input length is set to 512 for all models in PlainQAFact.The QA model (i.e., \texttt{Electra-large}) used in our \factscore is directly downloaded from QAFactEval \cite{fabbri-etal-2022-qafacteval}. However, we set its maximum input length to 512 (from 364) tokens to incorporate more source context. Similarly, to ensure that the answers extracted during the AE stage are valid (i.e., within the 512-token limit to maintain consistency with the subsequent QA model), we configure the LLMs' input lengths to 512 tokens. In MedCPT retrieval, we set \texttt{k} (the number of retrieved snippets) to 3, ensuring the retrieved information remains within a short context window for the subsequent QA process (\ref{appx:experiment-settings}). In the QF stage, we follow QAFactEval \cite{fabbri-etal-2022-qafacteval} and remove the unanswerable questions using the pre-trained \texttt{Electra-large} QA model\footnote{Download here: \url{https://github.com/salesforce/QAFactEval/blob/master/download_models.sh}}. 

We apply the default setting (i.e., \texttt{RoBERTa-base}) for AlignScore \cite{zha2023alignscore} and SummaC-Conv for SummaC \cite{laban2022summac}. For \texttt{GPT-4o} experiments as a factual consistency evaluator, we set the temperature as 1.0 in this paper. For the FareBio dataset, each plain language sentence is annotated with two binary labels: faithfulness and factual hallucination. In the released data, there are no sentences that are faithful but factually wrong. For the dataset pre-processing, we treat a sentence as not factual only when both labels are ``no.'' If either faithfulness or factual hallucination is ``yes,'' we treat the sentence as factual. To construct summary level labels, we first apply this rule to every sentence and then group sentences by their original FareBio summary. If a summary contains at least one sentence that is not factual, we label the whole summary as not factual, which produces 33 summaries. If all sentences in a summary are factual, we label the summary as factual, which produces 174 summaries. For the subset of 53 valid summaries used in our explanation focused evaluation, we further keep only those sentences that meet our filtering criteria for factual elaborative explanations and then aggregate them back into summaries.

For the computational costs of \texttt{GPT-4o}, it costs about \$0.01 per plain language summary during the factual consistency evaluation on the CELLS dataset experiment, combining input and output tokens, and takes about 2 seconds per summary. In contrast, \factscore takes about 30 seconds per summary on our hardware. However, \factscore is able to provide intermediate outputs for each step in the QA based evaluation, which makes our metric more transparent and easier to inspect than the black box \texttt{GPT-4o} evaluation, even in most cases where \texttt{GPT-4o} achieves slightly higher performance than \factscore.}

\subsection{\rev{LLM Prompts}} \label{appx:llm-prompt}

We utilize two types of prompts to guide LLMs through two stages of factual consistency evaluation: classification and answer extraction. In accordance with the benchmark annotation protocol (\ref{appx:annotation-protocol}), we employ \texttt{GPT-4o} as a classifier to determine whether a given sentence or summary requires elaborative explanations for factual verification. For both stages, we set the \texttt{max\_tokens} parameter to 512 and configure the temperature to 0 for \texttt{GPT-4o} and 0.01 for the \texttt{Llama 3.1} model.
\begin{lstlisting}[caption={Prompt of \texttt{GPT-4o} as the Classifier.},xleftmargin=0pt,xrightmargin=0pt]
@\textcolor{mygray}{Developer}@
Annotate whether a sentence or summary includes information not present in the original abstract. 

The sentence or summary contains external information that is not explicitly mentioned, paraphrased, or implied in the original abstract will be labeled as 'Yes'.

The sentence or summary contains information that is explicitly stated or closely paraphrased from the original abstract will be labeled as 'No'.

@\textcolor{mygray}{User}@
Sentence or summary: @\textcolor{myblue}{<input>}@
Original abstract: @\textcolor{mygreen}{<abstract>}@
\end{lstlisting}

For the AE stage, we explore both \texttt{GPT-4o} and \texttt{Llama 3.1} as backbone models to assess the factual consistency evaluation performance of \factscore. Following the task description outlined in QAFactEval \cite{fabbri-etal-2022-qafacteval}, we instruct both LLMs to extract potential answer entities from the input PLS.

\begin{lstlisting}[caption={Prompt of \texttt{GPT-4o} as the Answer Extractor.},xleftmargin=0pt,xrightmargin=0pt]
@\textcolor{mygray}{Developer}@
QA-based metrics compare information units between the summary and source, so it is thus necessary to first extract such units, or answers, from the given summary. Please extract answers or information units from a plain language summary.

@\textcolor{mygray}{User}@
Extract a comma-separated list of the most important keywords from the following text: @\textcolor{myblue}{<input>}@ 
\end{lstlisting}

\begin{lstlisting}[caption={Prompt of \texttt{Llama 3.1} as the Answer Extractor.},xleftmargin=0pt,xrightmargin=0pt]
@\textcolor{mygray}{System}@
QA-based metrics compare information units between the summary and source, so it is thus necessary to first extract such units, or answers, from the given summary. Please extract answers or information units from a plain language summary.

@\textcolor{mygray}{User}@
Extract a comma-separated list of the most important keywords from the following text: @\textcolor{myblue}{<input>}@ 
\end{lstlisting}

Additionally, we report the performance of using \texttt{Llama 3.1} as a judge to 
evaluate the factual consistency of a given PLS based on its source scientific abstract. 
\begin{lstlisting}[caption={Prompt of \texttt{Llama 3.1} as a factual consistency judge.},xleftmargin=0pt,xrightmargin=0pt]
@\textcolor{mygray}{System}@
Rate the factuality of the given plain language sentence or summary compared with the scientific abstract. Output a numeric score from 0 to 100, with 100 meaning the sentence is completely factually consistent with the abstract and 0 meaning the sentence is completely non-factual with the abstract.

@\textcolor{mygray}{User}@
Sentence or summary: @\textcolor{myblue}{<input>}@
Original abstract: @\textcolor{mygreen}{<abstract>}@
Factuality score (only output a numeric score): @\textcolor{myred}{<score>}@
\end{lstlisting}

For the LLM-based perturbation of \benchmark and CELLS datasets, we follow the protocol of APPLS \cite{guo-etal-2024-appls} on faithfulness criteria. 
\begin{lstlisting}[caption={Prompt of \texttt{GPT-4o} for faithfulness perturbation},xleftmargin=0pt,xrightmargin=0pt]
@\textcolor{mygray}{System}@
You are a data transformation assistant. You will receive a sentence from a biomedical literature. You will generate a new version of the given sentence based on the following rules for faithfulness perturbations:

1. Number Swap
Locate any numeric value(s) in the sentence and swap them with different numeric value(s).
Example: "infected more than 59 million people" -> "infected more than 64 million people"

2. Entity Swap
Locate a key entity (e.g., virus name, drug name, organization) in the sentence and swap it with a different entity.
Example: "coronavirus 2 (SARS-CoV-2)" -> "canine adenovirus (CAV-2)"

3. Synonym Verb Swap
Identify a key verb in the sentence and replace it with a near-synonym or related verb that changes the nuance or meaning slightly.
Example: "killed more than one of them" -> "stamped out more than one of them"

4. Hypernym/Antonym Swap
Select a word and replace it with either a hypernym (a more general term) or an antonym (opposite meaning), as appropriate.
Example (antonym): "killed more than one of them" -> "saved more than one of them"
Example (hypernym): "dog" -> "animal" (if relevant)

5. Negation
Negate a key part of the sentence to flip its meaning.
Example: "has infected more than 59 million people" -> "hasn't infected more than 59 million people" or "has not infected more than 59 million people"

Your task:
Read each sentence, generate a new sentence based on one of the five types of perturbation stratigies (Number Swap, Entity Swap, Synonym Verb Swap, Hypernym/Antonym Swap, Negation) above.
Return the perturbation sentence. 

Do not change the rest parts of the sentence except for the perturbation content. For example, the original sentence is "The skin patch and the vaginal (birth canal) ring are two methods of birth control." The perturbation sentence should be "The skin patch and the vaginal (birth canal) ring are five methods of birth control."

@\textcolor{mygray}{User}@
Sentence: @\textcolor{myblue}{<input>}@
Perturbation sentence: @\textcolor{myred}{<sentence>}@
\end{lstlisting}

\section{Pilot Study on FactPICO} \label{app:pilot-study}
To investigate the performance of existing automatic factual consistency evaluation metrics on plain language generation tasks, we employ the recently introduced FactPICO dataset \cite{joseph-etal-2024-factpico}. This dataset comprises human-labeled plain language summaries of Randomized Controlled Trials (RCTs) that address several critical elements: Populations, Interventions, Comparators, Outcomes (PICO), and additional information. All summaries are generated by various LLMs based on medical literature and include added information (i.e., extensive explanations) not present in the original abstracts. We hypothesize that existing factuality evaluation metrics for text summarization may struggle to accurately assess the factuality of this added information. To validate this assumption, we conduct pilot studies using four factual consistency evaluation metrics from FactPICO alongside one NLI-based metric: DAE \cite{goyal2021annotating}, AlignScore \cite{zha2023alignscore}, SummaC \cite{laban2022summac}, QAFactEval \cite{fabbri-etal-2022-qafacteval}, and QuestEval \cite{scialom-etal-2021-questeval}.

\subsection{Dataset Pre-processing}
FactPICO provides span-level annotations for LLM-generated summaries, assessing whether the additional information is present and determining its factuality, labeled as either ``yes'' or ``no.'' We first remove all special identification tags from the abstracts, such as ``ABSTRACT.BACKGROUND,'' ``ABSTRACT.RESULTS,'' and ``ABSTRACT.CONCLUSIONS.'' We then deduplicate the generated summaries, while retaining duplicates within abstracts, as each abstract has three summaries generated by different LLMs, and each abstract may have varying numbers of generated summaries, from none to multiple. This pilot study focuses on two key research questions: evaluating the effectiveness of existing factuality metrics in assessing added information and non-factual added information.

\subsection{Experiments and Results}
We conduct experiments using the processed FactPICO dataset to conduct pilot studies that assess the performance of existing factual consistency evaluation metrics in detecting added information in plain language summarization tasks. It is important to note that some outliers in the plain language summaries of the FactPICO dataset contain extraneous content, which causes the input lengths to exceed the limitations of both the DAE and QAFactEval metrics. Furthermore, the factuality of the LLM-generated plain language summaries is not guaranteed.

\begin{figure}[!h]
  \centering
  \includegraphics[width=0.9\textwidth]{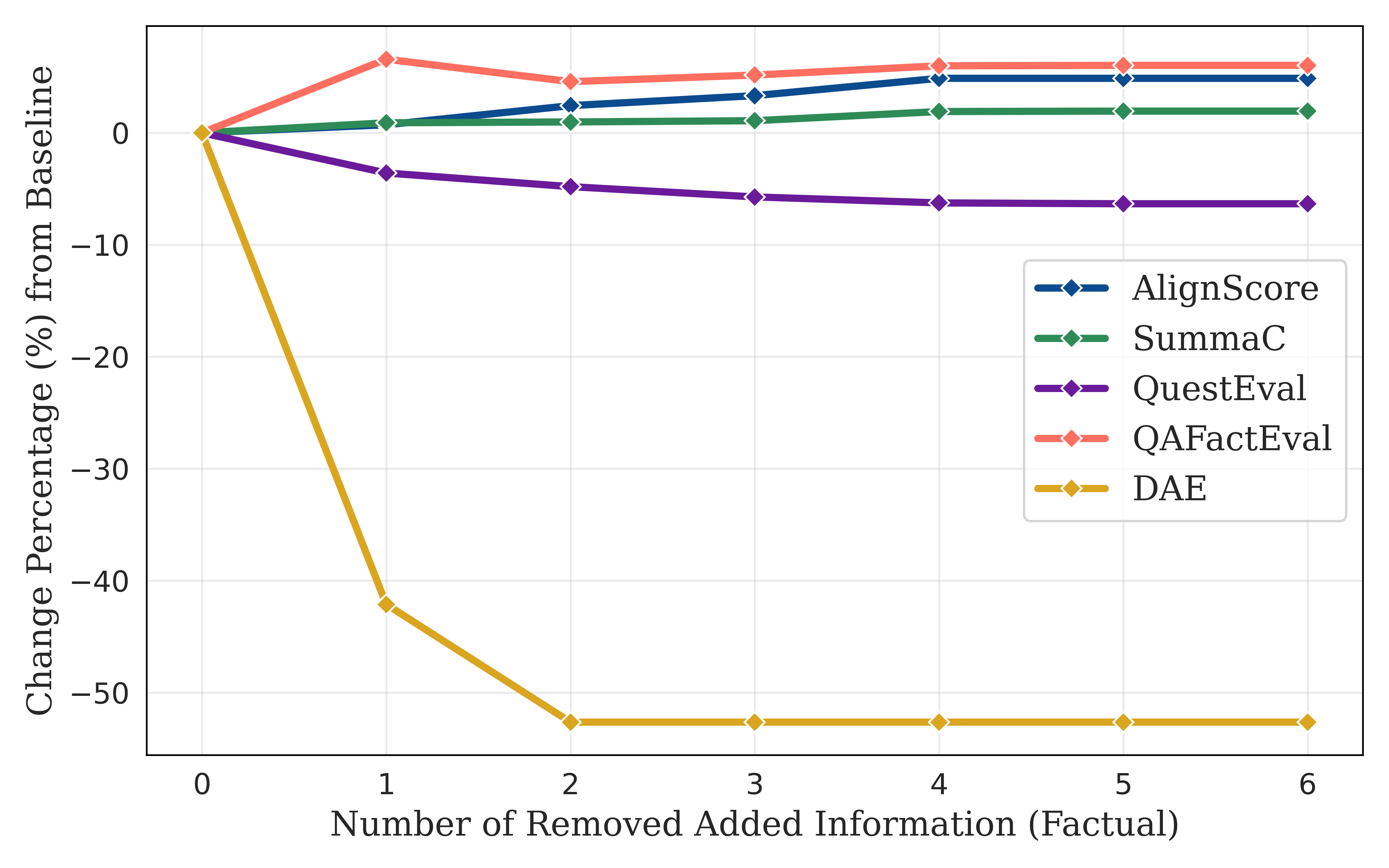}
  \caption{Score change percentage from baselines over five metrics on the FactPICO dataset in removing factual added information. We expect each metric stays unchanged even when more added factual information is removed. The evaluation dataset contains 88 valid summary-abstract pairs.}
  \label{fig:factpico-rq1}
\end{figure}

\begin{figure}[!h]
  \centering
  \includegraphics[width=0.9\textwidth]{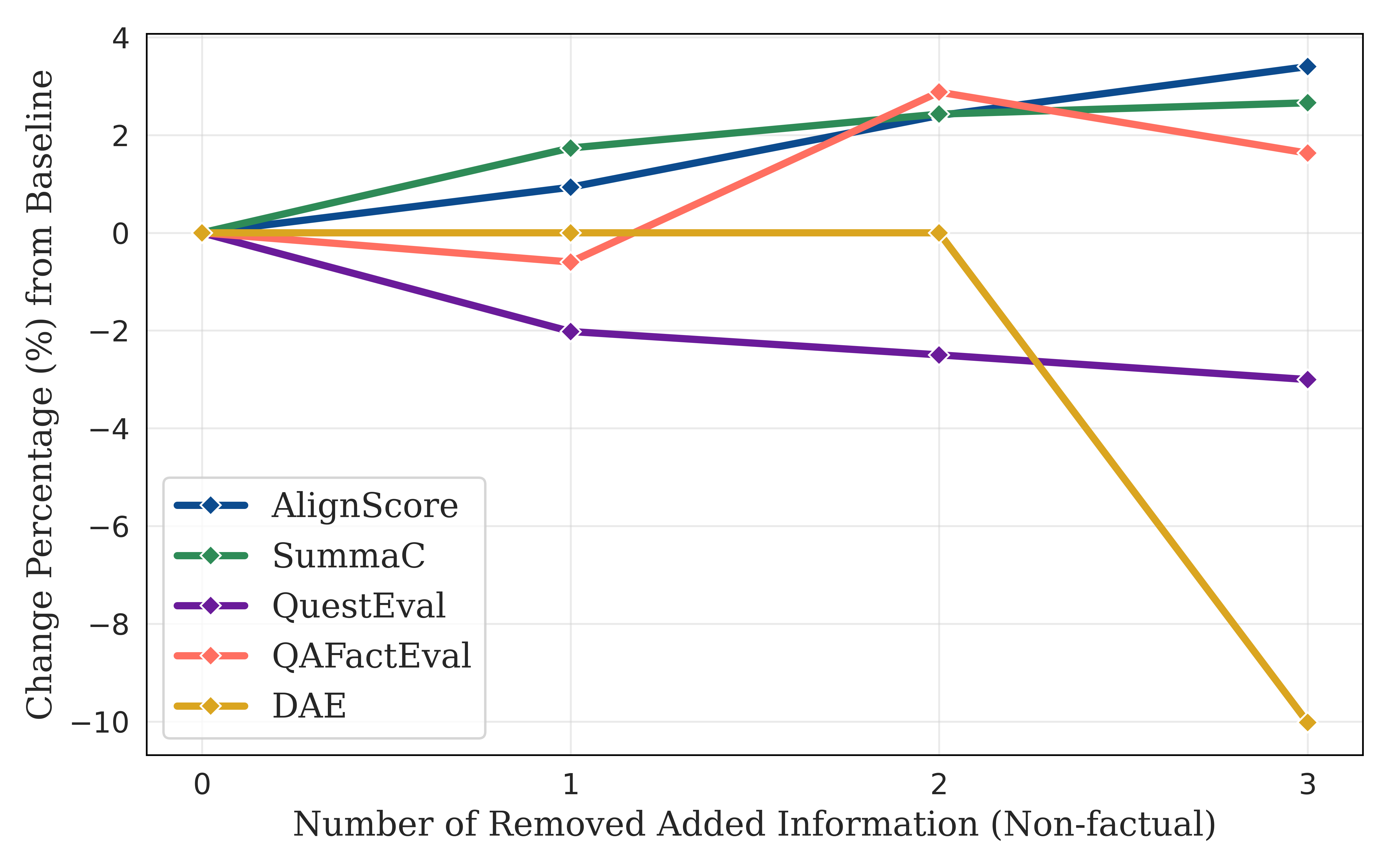}
  \caption{Score change percentage from baselines over five metrics on the FactPICO dataset in removing non-factual added information (60 pairs). We expect the change percentage from baseline increases when more added non-factual information is removed.}
  \label{fig:factpico-rq2}
\end{figure}

\textbf{RQ1: Do existing metrics perform well when external information is added to plain language summaries compared with no added information?}

In this study, we evaluate the ability of existing factual consistency evaluation metrics to detect added information in plain language summaries. We focus on summaries where all added information is annotated as factual, resulting in a dataset of 88 summary-abstract pairs. To assess metric sensitivity to the added information, we iteratively remove sentences from each plain language summary that contain added spans. For example, if a span such as ``of a medicine called haloperidol'' is labeled as factual (``yes''), we remove the entire sentence in the original plain language summaries containing that span through exact matching, continuing this process until no added information remains. This procedure enables us to determine how effectively current metrics handle added information that is absent from the original abstracts. As shown in Figure~\ref{fig:factpico-rq1}, we report performance changes relative to baseline scores. For example, the AlignScore evaluation increases by 3.3\% (on a 100-point scale) when six spans of added information are removed, resulting in an overall change of approximately 4.9\% compared to the setting in which no added information is removed. These findings indicate that added information affects model-based factuality evaluation metrics such as QAFactEval (6.0\%), AlignScore (4.9\%), and SummaC (1.9\%), with scores improving as more added information is removed. In contrast, both QuestEval and DAE scores decline with the removal of added information, and notably, DAE exhibits the most significant performance drop. Overall, these findings suggest that all the five metrics have difficulty accurately assessing added factual information, as evidenced by both increases and decreases in their performance.

\textbf{RQ2: Can existing metrics distinguish between non-factual and factual added information?}

In this research question, our goal is to evaluate the sensitivity of factuality evaluation metrics in detecting non-factual information within plain language summaries. The FactPICO dataset labels added information as either ``yes'' (factual) or ``no'' (non-factual). In this scenario, we sample only those plain language summaries that contain both ``yes'' and ``no'' labels for added spans. As with RQ1, we iteratively remove sentences containing non-factual added spans from each plain language summary until no non-factual sentences remain. Overall, as illustrated in Figure~\ref{fig:factpico-rq2}, only AlignScore, SummaC, and QAFactEval show improved performance as more non-factual information is removed, indicating that these metrics are sensitive in added non-factual information. Nevertheless, based on the results of RQ 1 and 2, our findings suggest that existing factual consistency evaluation metrics have limited capacity to accurately distinguish between factual and non-factual added information in plain language summarization tasks.

\section{Statistical Testing} \label{appx:stat-tests}
Following SummaC \cite{laban2022summac} and QAFactEval \cite{fabbri-etal-2022-qafacteval}, we test whether our proposed metric achieves statistically significant improvements over other methods. Given Kendall's~$\tau$ and Pearson correlations do not show the discriminative feature of continuous scores in various evaluation metric, we run a systematic evaluation on four datasets (CELLS, FareBio, FactPICO, and \benchmark) through bootstrap resampling \cite{efron1982jackknife} on the AUC-ROC results. We compare our best metric to other methods using the confidence intervals as significance level of \orange{0.01} and apply Bonferroni correction \cite{bonferroni1935calcolo} following SummaC \cite{laban2022summac}. Our results show that \factscore achieves statistically significant improvements over QuestEval \cite{scialom-etal-2021-questeval}, SummaC \cite{laban2022summac}, and DAE \cite{goyal2021annotating} on the CELLS and \benchmark datasets. As shown in Figure~\ref{fig:external-only}, \factscore also outperforms other automatic metrics in evaluating elaborative explanation sentences, except for QAFactEval \cite{fabbri-etal-2022-qafacteval}. This highlights the effectiveness of \factscore for biomedical PLS tasks that include external explanations. 

\rev{For hypothesis testing, we treat AUC-ROC as the primary metric. For each baseline $b$ we conduct a paired bootstrap test on the difference in AUC-ROC, $\Delta^{(k)} = \mathrm{AUC}^{(k)}_{\text{\factscore}} - \mathrm{AUC}^{(k)}_{b}$ for bootstrap replicate $k$. The one-sided $p$-value for the alternative ``\factscore{} $>$ baseline'' is estimated as $p = \Pr(\Delta \le 0)$ over the 10{,}000 replicates. Across the seven baselines we apply Holm-Bonferroni \cite{holm1979simple} correction to control the family-wise error rate at $\alpha = 0.01$.}

\section{\rev{Entity Swap Perturbation}} \label{appx:entity-swap}
\begin{table}[!ht]
\footnotesize
% \color{blue}
\begin{tabular*}{\textwidth}{@{\extracolsep\fill}llccc}
\toprule
\textbf{Dataset} & \textbf{Metrics} & \textbf{Kendall's~$\tau$} & \textbf{Pearson} & \textbf{AUC (99\% CI)} \\
\midrule
% \multicolumn{4}{l}{\textbf{CELLS}} \\
\multirow{8}{*}{CELLS} &    Llama 3.1      & 24.6 & 29.9 & 65.7* [57.8, 73.3] \\
    &    GPT-4o     & 69.3 & 74.3 & 98.4 [96.4, 99.7] \\
    \cmidrule(lr){2-5}
    & QAFactEval & 55.9 & 64.6 & 89.4 [85.5, 93.2] \\
    &    QuestEval  & 22.6 & 27.0 & 66.0* [58.7, 72.8] \\
    &    SummaC     & \cellcolor[HTML]{F4CCCC}16.4 & 24.5 & 61.7* [54.2, 68.7]\\
    &   AlignScore & 36.7 & 44.6 & 75.9* [69.5, 81.8]\\
    &    DAE        & 20.6 & \cellcolor[HTML]{F4CCCC}20.6 & \cellcolor[HTML]{F4CCCC}56.8* [52.6, 60.9] \\
    &    \factscore & \cellcolor[HTML]{D4EAD1}56.9 & \cellcolor[HTML]{D4EAD1}68.2 & \cellcolor[HTML]{D4EAD1}90.2 [85.9, 93.9] \\
\bottomrule
\end{tabular*}
\caption{Evaluation results of automatic metrics on the CELLS dataset \cite{guo2022cells} with entity swap non-factual plain language summary perturbation. Results are evaluated using Kendall’s~$\tau$, Pearson correlation, and AUC (AUC-ROC). * indicates a statistically significant difference compared to \factscore ($p < 0.01$). The standard deviations (std.) over five runs are: 0.1 (\factscore), 0.2 (\texttt{Llama 3.1}), and 2.5 (\texttt{GPT-4o}).}
\label{tab:entity-swap-results}
\end{table}

\begin{table}[!ht]
\footnotesize
% \color{blue}
\begin{tabular*}{\textwidth}{@{\extracolsep\fill}llcccc}
\toprule
\textbf{Dataset: CELLS}                 & \textbf{AE} & \textbf{Kendall's~$\tau$} & \textbf{Pearson} & \textbf{AUC (99\% CI)} & \textbf{std.} \\
\midrule
\multirow{2}{*}{w/ Heuristic Entity Swap} 
                                   & Llama 3.1         & \textbf{56.9} &\textbf{68.2}   & \textbf{90.2} [85.9, 93.9] & 0.1   \\
                                   & PromptRank & 49.6  & 58.8  & 85.0* [79.7, 89.6] & -  \\
\midrule
\multirow{2}{*}{w/ LLM Perturbation} 
                                   & Llama 3.1               & \textbf{62.1} & \textbf{75.2}   & \textbf{93.8} [90.3, 96.7] & 0.1   \\
                                   & PromptRank            & 56.3 &  66.6  &  89.6* [85.3, 93.3]  &  -  \\
\bottomrule
\end{tabular*}
\caption{Ablation study of answer extraction approach in the \factscore on the CELLS dataset. w/ Heuristic Entity Swap uses APPLS entity swap \cite{guo-etal-2024-appls} approach to create non-factual plain language summaries, while w/ LLM Perturbation applies \texttt{GPT-4o} to perturb the plain language summaries. AE = answer extraction. AUC = AUC-ROC. * indicates a statistically significant difference compared to \texttt{Llama 3.1} ($p < 0.01$). We run each setting for five times and report the standard deviation (std.). Note that only the settings using LLMs may produce fluctuated scores.}
\label{tab:answer-extractor-ablation}
\end{table}

\rev{As we discussed in Section~\ref{sec:main_result}, in addition to evaluating the performance of automatic factual consistency evaluation metrics on LLM-perturbed datasets, we also use the non-LLM entity swap approach introduced in \citet{guo-etal-2024-appls} to assess the robustness of our proposed metric. Specifically, we perturb plain language summaries using the CELLS dataset following the same entity swap protocol as APPLS \cite{guo-etal-2024-appls}. The entity swap system identifies specific ``entity spans'' in the text and substitutes them with existing terms from the UMLS knowledge base. As shown in Table~\ref{tab:entity-swap-results}, \factscore achieves the best performance among all automatic evaluation metrics on the CELLS dataset. \factscore achieves the highest correlation with human judgement (AUC=90.2, 99\% CI [85.9, 93.9]). As comparative effect sizes, we report pairwise differences between \factscore and the strongest baseline QAFactEval. The $\Delta\mathrm{AUC} = 0.7$, where $\Delta$ is \factscore minus the baseline. After correction, \factscore shows significantly higher AUC than \texttt{Llama 3.1}, QuestEval, SummaC, AlignScore, and DAE ($p < 0.01$, Holm-Bonferroni corrected for all), but is not significantly different from QAFactEval ($p = 0.64$) or \texttt{GPT-4o} ($p = 1.0$).

We also report an ablation study using a non-LLM answer extractor to show the effectiveness of our LLM-powered answer extraction approach. In Table~\ref{tab:answer-extractor-ablation}, our answer extraction method using \texttt{Llama 3.1} consistently achieves higher correlation (AUC=90.2, 99\% CI [85.9, 93.9] in w/ heuristic entity swap and AUC=93.8, 99\% CI [90.3, 96.7] in w/ LLM perturbation) compared with the non-LLM answer extractor (i.e., PromptRank). Using \texttt{Llama 3.1} shows significantly higher AUC than PromptRank in both settings ($p < 0.01$, Holm-Bonferroni corrected for all), highlighting the robustness and effectiveness of LLM-powered answer extraction in factual consistency evaluation.}

\section{Summary-Level Explanation-Only Evaluation} \label{app:explanation-only}

\begin{figure*}[!h]
  \centering
  \includegraphics[width=0.9\textwidth]{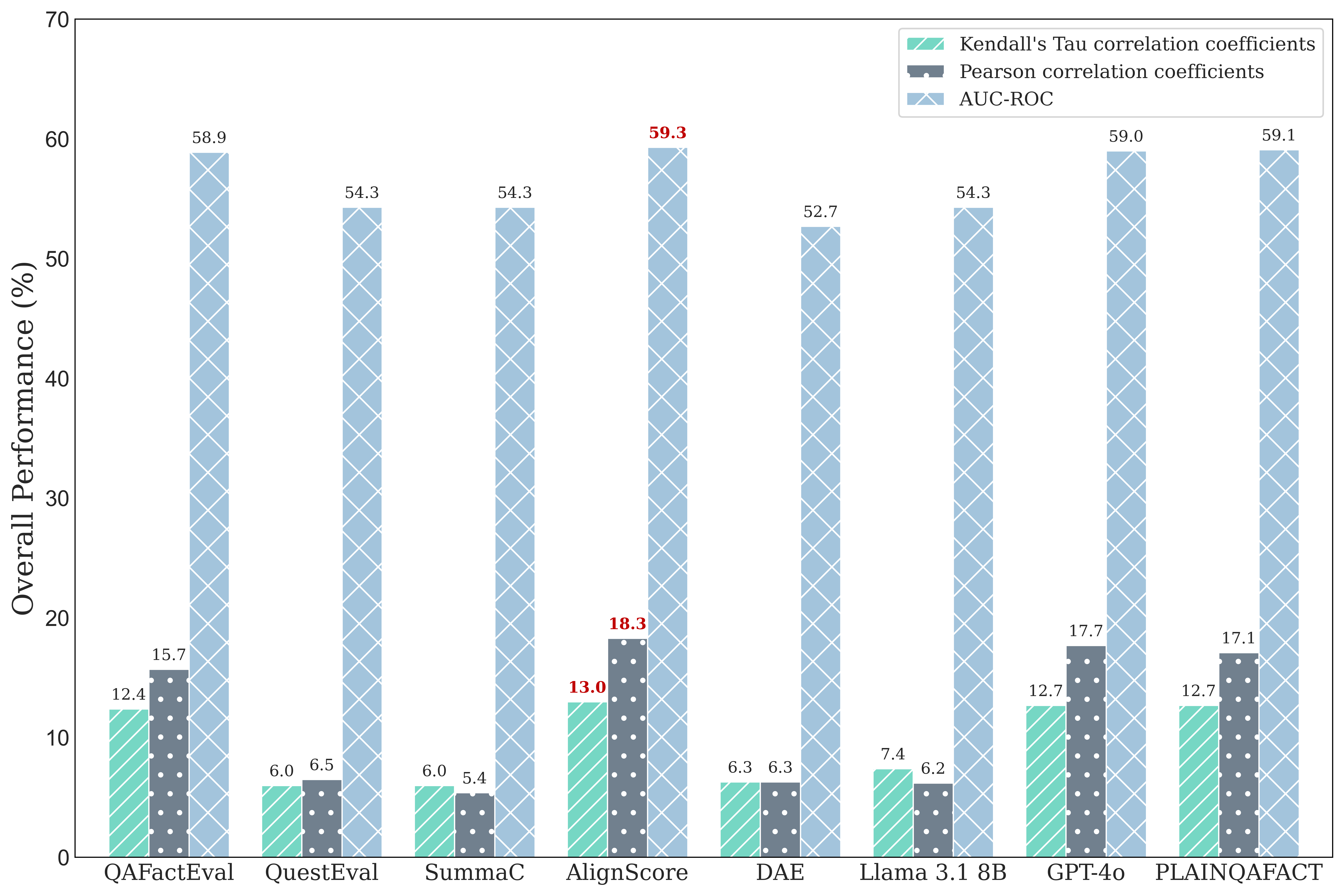}
  \caption{Overall performance on summaries containing added information (i.e., elaborative explanations) from FactPICO \cite{joseph-etal-2024-factpico}. The std. of \factscore, \texttt{Llama 3.1}, and \texttt{GPT-4o} are 0.2, 0.2, and 3.3, respectively, based on five runs of each metric.}
  \label{fig:factpico-external-only}
\end{figure*}

\begin{figure*}[!h]
  \centering
  \includegraphics[width=0.9\textwidth]{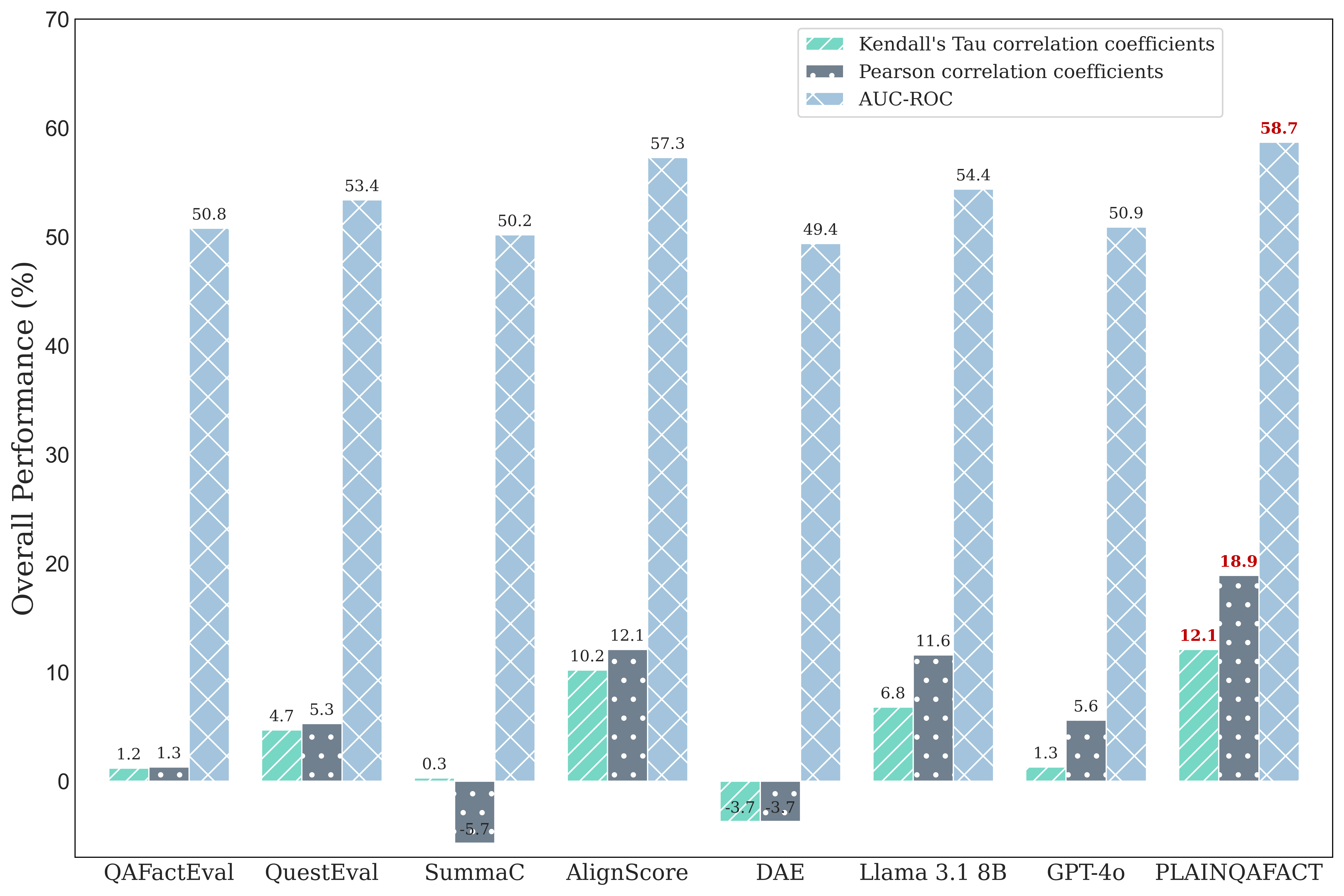}
  \caption{Overall performance on summaries with elaborative explanations from FareBio \cite{fang2024understanding}. The std. for \factscore, \texttt{Llama 3.1}, and \texttt{GPT-4o} are 0.03, 0.3, and 5.7, respectively, computed over five runs.}
  \label{fig:farebio-external-only}
\end{figure*}

Similar to our sentence-level explanation annotations in \benchmark, FactPICO \cite{joseph-etal-2024-factpico} and FareBio \cite{fang2024understanding} datasets also provide human-annotated explanation information in plain language summaries. According to the results shown in Section~\ref{sec:explaination_result}, we also assess the performance of five factual consistency evaluation metrics under FactPICO and FareBio datasets.

In the FactPICO, the elaborative explanation is defined as ``added information.'' We first select summaries that include added information, and then we label a summary as ``non-factual'' if it contains any non-factual added information as determined by annotators. We generate a summary-level explanation-only FactPICO dataset consisting of 231 summary-abstract pairs. As shown in Figure~\ref{fig:factpico-external-only}, AlignScore achieves the best performance across all three evaluation criteria. However, since the summaries in the FactPICO dataset are generated using LLMs, the factual consistency between the generated summaries and the original source abstracts is not guaranteed. For instance, we treat a summary as ``factual'' indicates that all the added information is factual, but it does not necessarily reflect the factual consistency of other content. Therefore, the results presented in Figure~\ref{fig:factpico-external-only} may become biased.

For FareBio \cite{fang2024understanding}, we collect explanation sentences based on both ``faithfulness'' and ``factual hallucination'' labels in the original dataset. As shown in Figure~\ref{fig:farebio-external-only}, \factscore achieves the best performance compared with all other metrics, showing a clear advantage over \texttt{GPT-4o}. Since the FareBio dataset provides sentence-level annotations of factual consistency for each sentence in the plain language summaries, the results are more reliable than those from FactPICO. \rev{On both datasets, we do not find statistically significant differences in AUC-ROC metric by $p<0.01$. While some existing metrics perform similarly in certain settings, our metric consistently shows better performance in correlation. }

\section{\orange{Quantitative Error Analysis of FareBio}} \label{app:quantative-error-analysis}
\orange{Given the Table~\ref{tab:farebio-error-analysis}, we further conduct quantitative analysis regarding failure cases in FareBio using our proposed metric. We randomly sample 30 plain language summaries (both factual and non-factual) from FareBio and evaluate all sentences within them, resulting in 324 split sentences. We categorize errors into three main types:

1. QA Pipeline Errors (34.0\% of errors): misclassification of sentence types, retrieval of irrelevant context, and generation of unanswerable questions.

2. Answer Extraction and Evaluation Issues (18.4\% of errors): failure to extract numerical values and answer mismatches when using BERTScore.

3. Dataset Labeling Issues (47.7\% of errors): inconsistencies in FaithBio's annotation protocol.

Table~\ref{tab:farebio-quantitative-error} presents the complete quantitative breakdown. We identify 109 errors across the 324 sentences (33.6\% error rate). Notably, 47.7\% of errors occurs in sentences mislabeled as factual and non-factual, with most being false negatives. FareBio uses a different annotation protocol than \benchmark. All source-grounded simplification sentences are implicitly treated as factual. Annotators do not evaluate factuality once a sentence is labeled as faithful. This creates label noise when evaluating simplification sentences, as the ground truth may not reflect actual factual consistency. Our analysis reveals that \factscore performs better on \benchmark (where annotation protocols align with our metric's design) than on FareBio. The performance gap reflects both limitations in our approach, particularly for numerical verification, and incompatibilities between FareBio's annotation protocol and our QA components. }

\begin{table}[!ht]
\centering
\footnotesize
\setlength{\tabcolsep}{6pt}
\renewcommand{\arraystretch}{1.4}
% \color{orange}
\begin{tabular}{@{} l c c c c c @{}}
\toprule
\textbf{Error Type} & \textbf{Count} & \textbf{\% of Errors} & \textbf{\% of Total} & \textbf{In F} & \textbf{In NF} \\
\midrule
\multicolumn{6}{l}{\textit{\textbf{QA Pipeline Errors}}} \\
\quad Misclassification  & 13 & 11.9\% & 4.0\% & 10 & 3 \\
\quad Retrieval quality & 9 & 8.3\% & 2.8\% & 8 & 1 \\
\quad Unanswerable generated question & 15 & 13.8\% & 4.6\% & 11 & 4 \\
\midrule
\multicolumn{6}{l}{\textit{\textbf{Answer Extraction and Evaluation Issues}}} \\
\quad Numerical value mis-extraction & 16 & 14.7\% & 4.9\% & 11 & 5 \\
\quad Answer match misalignment & 4 & 3.7\% & 1.2\% & 3 & 1 \\
\midrule
\multicolumn{6}{l}{\textit{\textbf{Dataset Issues}}} \\
\quad False positive (F labeled as NF) & 10 & 9.2\% & 3.1\% & 0 & 10 \\
\quad False negative (NF labeled as F) & 42 & 38.5\% & 13.0\% & 42 & 0 \\
\midrule
\multicolumn{1}{l}{\textbf{Total Errors}} & 109 & 100.0\% & 33.6\% & 85 & 24 \\
\multicolumn{1}{l}{\textbf{Correct Classifications}} & 215 & -- & 66.4\% & 114 & 101 \\
\midrule
\multicolumn{1}{l}{\textbf{Total Sentences Analyzed}} & 324 & -- & 100.0\% & 198 & 126 \\
\bottomrule
\end{tabular}
\caption{Quantitative error analysis of \factscore on 60 randomly sampled summaries from FareBio dataset (30 factual, 30 non-factual). Summaries are split into sentences when evaluating the factual consistency. F = factual sentences; NF = non-factual sentences.}
\label{tab:farebio-quantitative-error}
\end{table}


\end{document}